\documentclass[10pt,times,twocolumn]{article}
\usepackage{epsfig}
\usepackage{sbcm2019}
\usepackage{graphicx,url}
\usepackage{dirtytalk}
\usepackage{amsmath}
\usepackage{tikz}
\usepackage{makecell}
\def\checkmark{\tikz\fill[scale=0.4](0,.35) -- (.25,0) -- (1,.7) -- (.25,.15) -- cycle;}
\usepackage{graphicx}

\usepackage{anonymize} 

 \usepackage[utf8]{inputenc}
 
\title{A Comprehensive Review on Summarizing Financial News Using Deep Learning}

\author{\anonymize{Saurabh Kamal}\inst{1} \and
        \anonymize{Sahil Sharma}\inst{2}}

\address{\anonymize{Engineering and Technology Department, Liverpool John Moores University, Liverpool, United Kingdom}
\nextinstitute
\anonymize{Computer Science and Engineering Department, Thapar Institute of Engineering and Technology, Patiala, India}
\email{\anonymize{saurabh.kamal1@gmail.com, sahil301290@gmail.com}}
}

\begin{document}

\maketitle

\begin{abstract}
Investors make investment decisions depending on several factors such as fundamental analysis, technical analysis, and quantitative analysis. Another factor on which investors can make investment decisions is through sentiment analysis of news headlines, the sole purpose of this study. Natural Language Processing techniques are typically used to deal with such a large amount of data and get valuable information out of it. NLP algorithms convert raw text into numerical representations that machines can easily understand and interpret. This conversion can be done using various embedding techniques. In this research, embedding techniques used are BoW, TF-IDF, Word2Vec, BERT, GloVe, and FastText, and then fed to deep learning models such as RNN and LSTM. This work aims to evaluate these models’ performance to choose the robust model in identifying the significant factors influencing the prediction. During this research, it was expected that Deep Learning would be applied to get the desired results or achieve better accuracy than the state-of-the-art. The models are compared to check their outputs to know which one has performed better.  
\end{abstract}

\section{Introduction}\label{sec:Intro}

Any investor’s objective is to predict the market behavior to make the best possible decision when he/she buys or sells stocks intending to make profits. It is a challenging task because investors’ sentiments are volatile and influenced by many factors such as politics, the global economy, investors’ expectations, etc. However, investor sentiment on the stock market is more evident in socially more prone to herd-like behavior, overreaction, and low institutional involvement. Through sentiment analysis, investors attempt to ascertain when the equity market is being driven by sentiments rather than rational decision-making. This research is conducted from the fundamental analyst’s perspective to analyze external information as political and economic factors. The details are extracted from unstructured data such as financial news articles or headlines. Many authors have suggested using text mining and machine learning techniques to examine the text and obtain information relevant to the forecast procedure. There have been several studies conducted in this field of Machine Learning or Deep Learning and Natural Language Processing (NLP) – “Extracting and Analyzing Sentiments from News Headlines in Correlation with a Stock Market or Predict Company’s Specific Stock on its News Sentiments.”

\section{Literature Review}

Stock Market prediction based on news headlines has 
always fascinated researchers. There have been numerous researches done on this topic, “how news headlines affect the stock price or how the International News impact the Major Global Indices?” For an investor, it is crucial to access headlines regularly. There are three kinds of news to which the financial market may or may not respond at all:-
\begin{itemize}
  \item Positive News.
  \item Negative News.
  \item Neutral News.
\end{itemize}
\noindent
A person has to be competent enough to make sense of the information and swiftly understand whether his/her stock will go up or down.
\begin{figure}[h]
\centering
  \includegraphics[width=.5\textwidth]{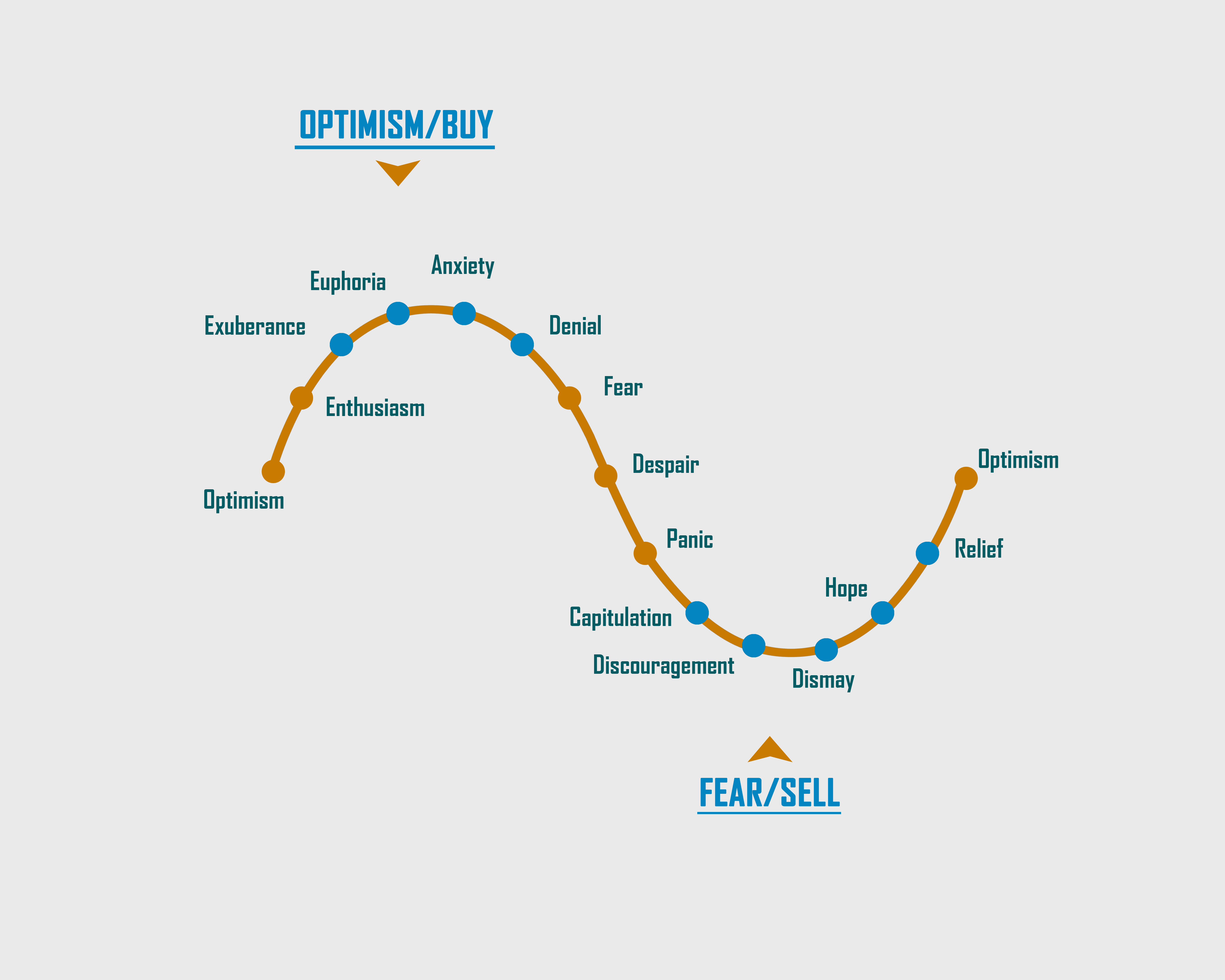}
\caption{The Cycle of Market Emotions}
\label{fig:exampleFig}
\end{figure}

\subsection{Good / Positive News}

Financial news considered positive or good is liable to have an optimistic impact on the stock exchange, and one can understand how stock prices increase as the news article comes out. Positive information such as joint venture deals, obtaining new contracts, outstanding financial results of a business, unearthing massive oil reserves, substantial sales numbers, etc., should make a stock rise. Share prices respond gradually but progressively to the good news. However, good news doesn’t all the time transform into a jump in stock price. Optimistic news at home and depressed stories overseas can negatively push the value of the stock down. Both the global market and home market are interconnected with each other. Sometimes, a bit of bad news from overseas can put the stock market down for a day.

\subsection{Bad / Negative News}

Bad News has a more profound effect on equity prices and investors’ sentiments than good news. The emotions of the market are a crucial issue. A gloomy environment or a disturbing story is enough to send the stock tumbling. It can stop a typical person from investing in shares. Business news has an immediate influence on the financial markets. It can convert a worse day into a terrific one or an incredible day into a worse one. The next headline can turn out to be a boom-and-bust.
Terrific news will have an encouraging effect on stock prices, but not always. If the bad news across the Globe overshadows the good news, it can turn out to be a bad day for stocks since stock price reacts to bad headlines more sharply than it would respond to  good news.
\begin{figure}[h]
\centering
  \includegraphics[width=.5\textwidth]{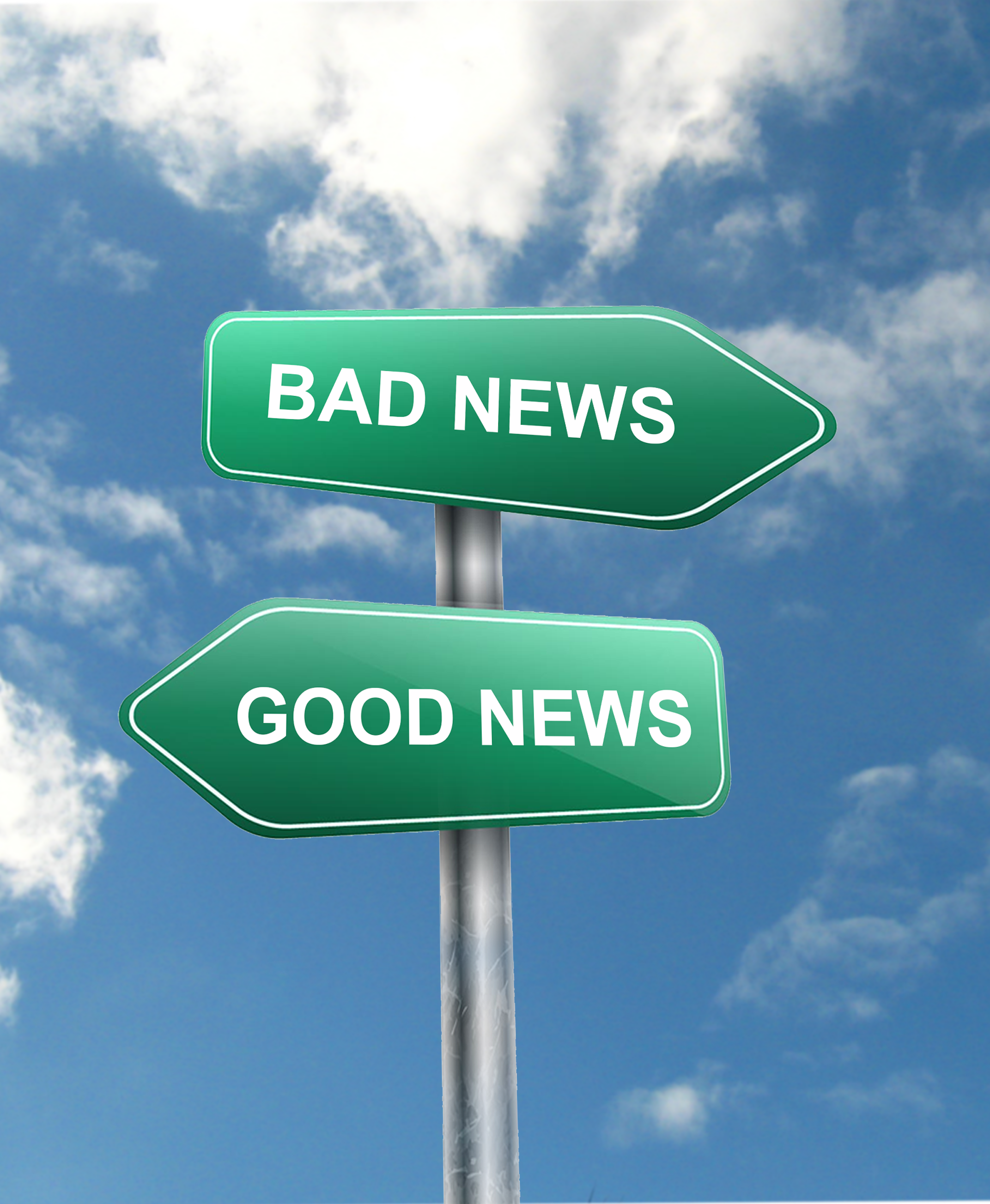}
\caption{Good News and Bad News}
\label{fig:exampleFig}
\end{figure}

\subsection{Interconnection between Sentiment and Stock Price}

In 2013, a phony tweet on Twitter caused the Dow Jones plunge \cite{APTwitter}

\noindent
\say{The three-minute plunge briefly wiped out USD 136.5 billion of the S and P 500’s value}
\begin{figure}[h]
\centering
  \includegraphics[width=.5\textwidth]{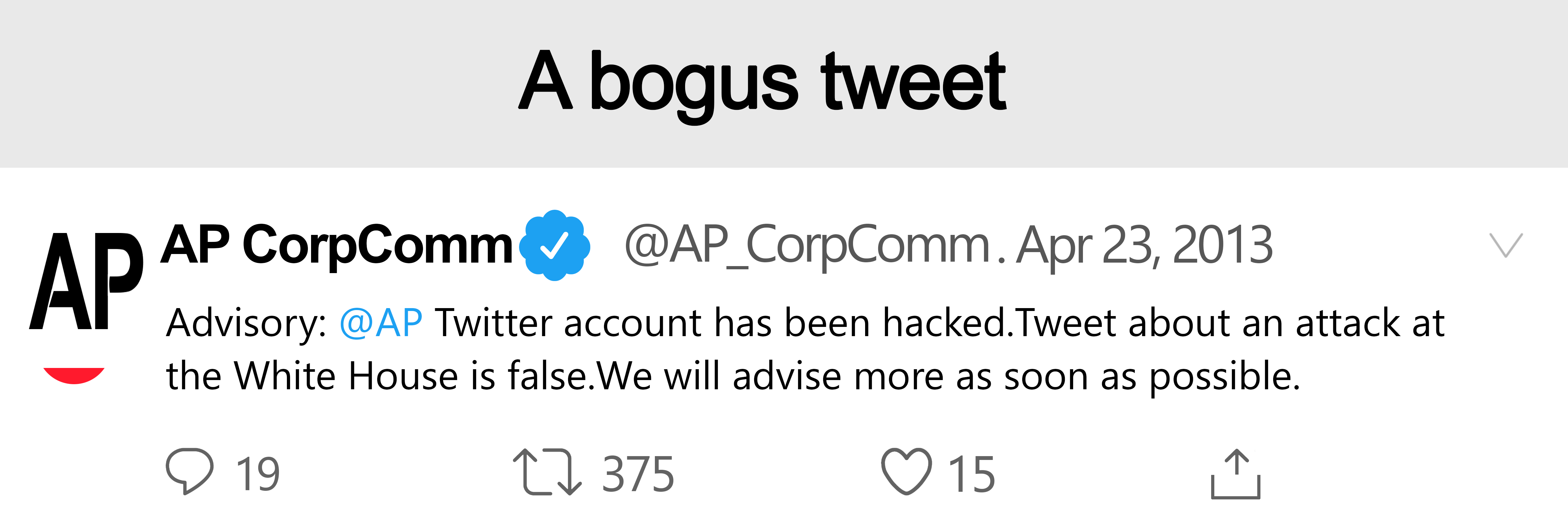}
\caption{A bogus tweet that caused Dow Jones collapse}
\label{fig:exampleFig}
\end{figure}
\noindent
After hacking the Associated Press’ Twitter account, a bogus tweet was posted that stated about the two blasts on the White House and that President Barak Obama got hurt.
\noindent
Once the tweet was disseminated to the market, the Standard and Poor’s 500 dropped by 14.6 points within 3 minutes – between 1:09 PM and 1:12 PM. The Dow 30 dropped more than 150 points, then regained quickly. However, keeping a tap on financial news is not an easy task since the internet is bombarded with lots of available information, and analyzing the sentiments (positive, negative, and neutral) of each headline, article, tweet, etc., is very tedious for every individual. Here is where machine learning techniques play a significant role in understanding whether a particular news article or headline is positive, negative, or neutral.
\noindent\\
Any machine learning model that expects to accomplish suitable precision needs to ascertain what verbal form is appropriate for the prediction, understand speech patterns, expressions, phrases, etc. and incorporate it into a reasonable decision on sentiments. This study’s primary purpose is to demonstrate and compare the performance of various deep learning models -  Recurrent Neural Network (RNN), and Long Short-Term Memory (LSTM) with word embedding techniques – Bag-of-Words (BoW), Term Frequency – Inverse Document Frequency (TF-IDF), Word2Vec, Bidirectional Encoder Representations from Transformers (BERT), Global Vectors for Word Representation (GloVe), and FastText \cite{FastTextRepresentation} to identify an accurate and efficient model. The study also understands how Global Financial News affects the stock market indices, not only on a National Scale but also on Global Level – understanding when positive sentiments lead to buying, when negative emotions lead to selling, and when to do nothing? How different country’s indices are related to each other? How are these indices associated with the financial News? Do national indices (Nifty100 or DAX30) get affected by the Global Financial News?

\subsection{Sentiment Analysis}

Sentiment  Analysis is one of the Natural Language Processing (NLP) subjects committed to discovering opinions or sentiments accumulated from multiple sources in a particular area. It is a method to detect and obtain views to gain an advantage in business, politics, media, public relations, etc. Sentiment Analysis, also known as opinion mining, is usually used to sense emotions in social data, understand customers, determine the label’s reputation, financial news related to the stock market, etc. and categorize their polarity into positive, negative, or neutral. In other words, sentiment analysis or opinion mining investigates the customers, audience, or investors’ mindset, classifies the text, and maps it to a class. It can be binary (positive or negative) or multi-class with three or more levels involved. Sentiment Analysis is fundamentally a classification algorithm intended to search opinionated points of view and accentuate certain significant information. “Opinions” in sentiment analysis is an impressionistic analysis of something established on empirical experience. To a certain extent, “opinions” are partly ingrained in facts and partially directed by sentiments. This research paper \cite{Samuels2019SentimentReviews} presents a relative sentiment analysis of extracted reviews from amazon.com of numerous smartphone opinions divided into positive, negative, and neutral. Every opinion was first organized into labeled words. Each word of the opinions has its part of speech. The POS tagging (Part-of-Speech) used in this study arranges phrases depending on their parts of speech.   Another area where sentiment analysis plays a crucial role is tweets from Twitter or posts from Facebook. On these platforms, most people voice their opinions or express their emotions towards varied topics. One such recent case is the COVID-19 pandemic. This paper \cite{Rajput2020WordPandemic} discusses the tweets’ computational analysis related to the COVID-19 epidemic from Jan 2020 to April 2020 and conducted two experimental studies, the first one on a word or phrase frequency and the second on distinctive tweet sentiments. For term frequency, unigram, bigram, and trigram are used by the power-law to capture the text analysis patterns. The power law is implemented in various findings. Mathematical representation of power law:
\begin{equation} \label{eq1}
m = k n^t
\end{equation}
\noindent
Where, m and n are variables of interest, t is the law’s exponent or scaling, and k is constant.
\\
\noindent\\
In this research work, the power-law models the term-frequencies of distinctive tweets. To evaluate the model, three metrics – Sum of squared estimate of errors (SSE), R-Squared, and Root Mean Squared Error (RMSE) – are being used. R-Squared in all three cases showed the best result (Unigram-0.9172, Bigram-0.8718, Trigram-0.9461) than that of SSE and RMSE that were relatively low. The Unigram word cloud showed Coronavirus as the most recurrent word. This paper uses TextBlob for sentiment analysis of tweets relevant to the Corona Virus Disease (COVID-19) epidemic for individuals’ separate tweets. TextBlob python library is used to understand the Global Political Environment \cite{UnderstandingScienceb}. Over the previous ten years, the usage of Twitter has risen steeply. These days it seems every politician feels the  urge to post his or her thoughts. Several political leaders have adopted Twitter as their principal mode of announcement to the public. So, it has got us to think that could we gauge how Global leaders interact on Twitter?
\begin{figure}[h]
\centering
  \includegraphics[width=.5\textwidth]{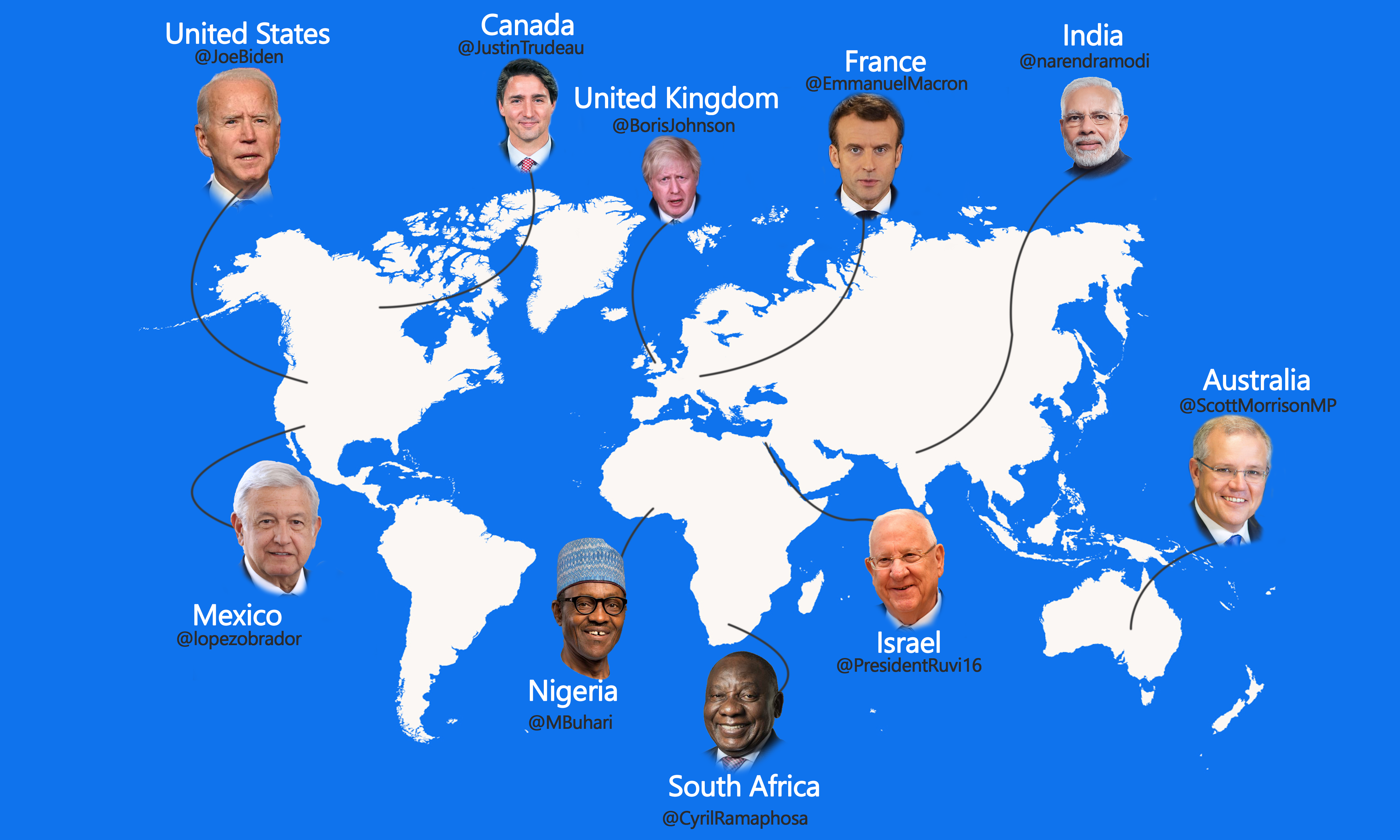}
\caption{Politician’s Twitter Accounts}
\label{fig:exampleFig}
\end{figure}
\cite{10Data} gathered approximately 1000 tweets citing two verifiable Twitter accounts, @vijayrupanibjp, and @BharatSolankee, chiefs of Gujarat Legislative Assembly election, 2017. This research executed sentiment analysis with SYUZHET in the CRAN package formulated on NRC Emotion Lexicon by using R programming. The following Figures demonstrate Vijay Rupani and Bharatsinh Madhavsinh Solanki’s sentiments throughout the Gujarat Assembly election of 2017.
\begin{figure}[h]
\centering
  \includegraphics[width=.5\textwidth]{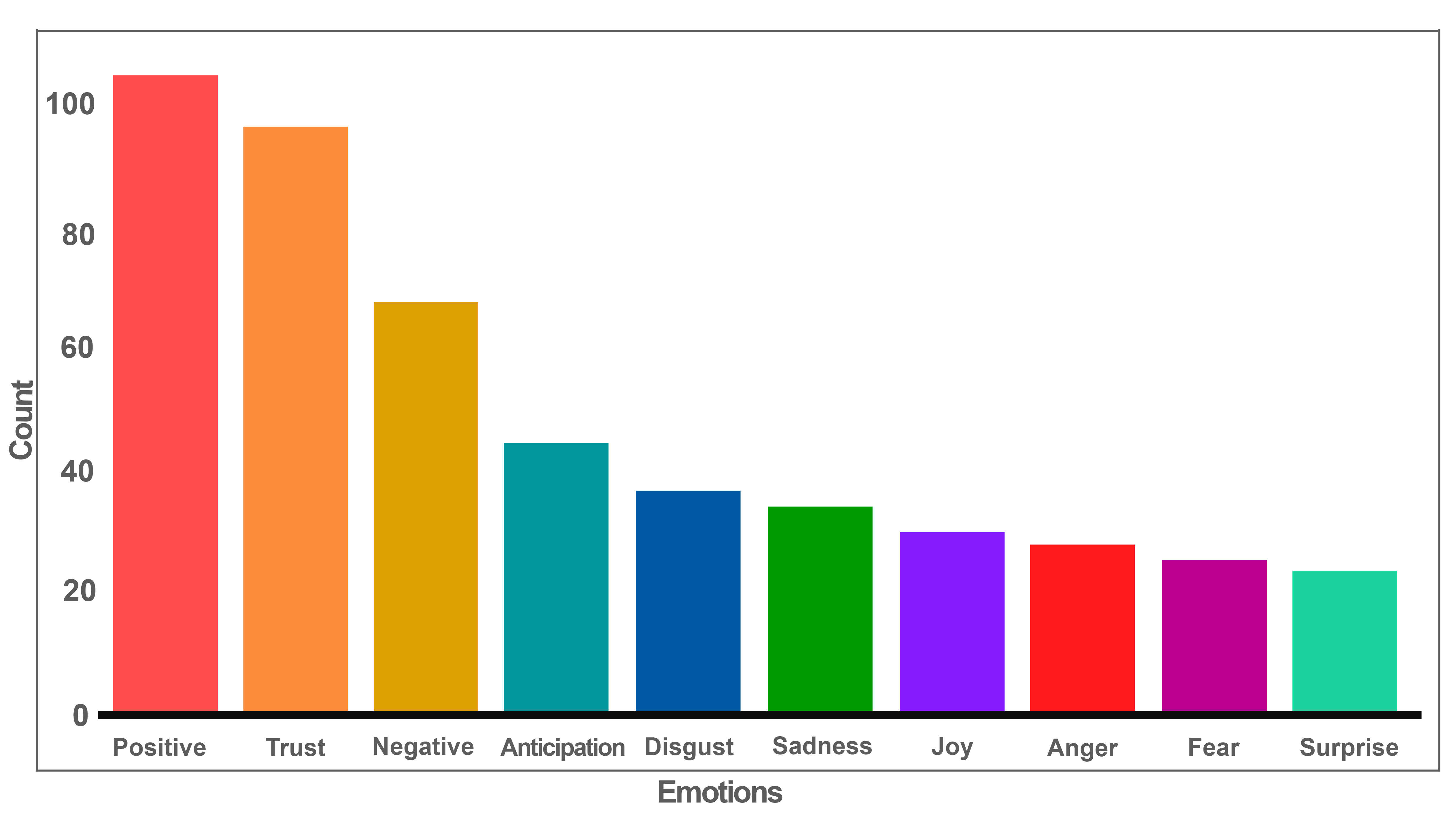}
\caption{Sentiments of Vijay Rupani during Gujarat Assembly elections, 2017}
\label{fig:exampleFig}
\end{figure}
Fig. 5 shows more than 300 tweets as positive sentiments for Vijay Rupani (BJP), whereas Fig. 6 offers more than 100 tweets as positive sentiments for Bharatsinh Madhavsinh. The Parallel Dots AI API calculated the sentiment score, distributing positive, negative, and neutral emotions for the event. The estimation later unveiled that 55\texttt{\%} of positive tweets for Vijay Rupani (BJP) have the probability of winning the General Assembly elections.
\begin{figure}[h]
\centering
  \includegraphics[width=.5\textwidth]{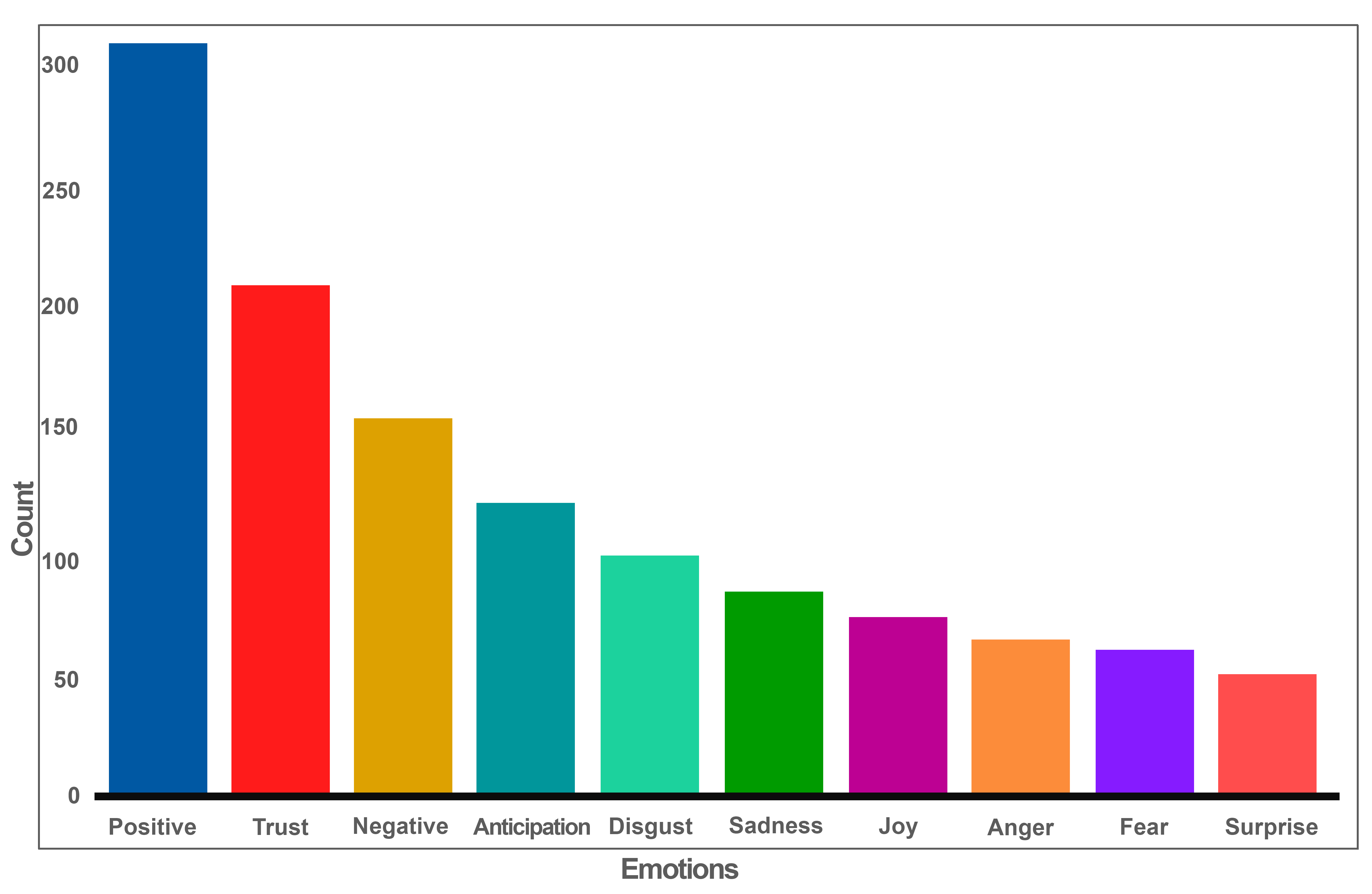}
\caption{Sentiments of Bharatsinh Madhavsinh during Gujarat Assembly elections, 2017}
\label{fig:exampleFig}
\end{figure}
\noindent
If it comes to tweets, then who can forget Donald J. Trump’s famous tweets. It enticed global attention. The President often took Twitter to make remarks about other world leaders, celebrities, citizens, and media channels. From his formal announcement of candidacy until his term, he tweeted more than 17,000 times \cite{Statista}. This paper \cite{Sahu2020SupervisedTechnique} examines the correlation between the President’s tweets and his acceptance ratings by applying sentiment analysis and data visualization techniques. The tweets are mined and given sentiment scores. A causative relationship between President’s Twitter activities and his approval ratings was found by employing cross-correlation analysis.  About a five-day interval was found by applying cross-correlation between a shift in the President’s approval rating concerning his twitter endeavors. Here, the unique problem was to compute the sentiment score for every tweet posted by then President Donald J.Trump (@RealDonaldTrump). To calculate the polarity and subjectivity, TextBlob was used. NLP techniques such as N-grams, Bag-of-words (BoW), and Term Frequency – Inverse Document Frequency (TF-IDF) were used for mining the features from the textual data. Another area where sentiment analysis has become vital is in determining critics’ thoughts concerning their comments. Analysis can rate how a movie review can be positive or negative. Hence, whether a movie review is positive or negative can be understood once the machine assimilates through training and testing the data. This \cite{Sahu2016SentimentAlgorithms} paper studies the emotions given in the IMDB movie review database’s opinions. The author had collected 50,000 reviews with the utmost 30 reviews for each movie. It analyzes the emotions to arrange the movie review’s polarity on a scale of 0 to 4, with 0 being highly disliked and four being highly liked. The computational method used in this paper is based on the publicly available library SentiWordNet. For determining the document’s polarity, this paper has focused on two areas:
\begin{enumerate}
   \item Feature selection - pre-processing, feature extraction, and feature reduction, and another is ranking. For pre-processing and feature extraction, POS tagging, SentiWordNet, and N-grams are applied to this research.
   \item Classification techniques - explicitly bagging, random forest, Naïve Bayes, k-nearest neighbor, and regression. Among these machine learning techniques used in this paper, random forest performs best with an accuracy of 88.95\texttt{\%}, and Naïve Bayes performs worst with an accuracy of 54.77\texttt{\%}.
\end{enumerate}
\noindent
Online reviews are worthwhile both for consumers and for retailers to make decisions while purchasing and to improve the quality of their products. This paper \cite{Fauzi2019Word2VecLanguage}, discovers the usage of the Word2Vec model with Support Vector Machine (SVM) for the classification tasks. There were a total of 772 reviews extricated from FemaleDaily.com. 50\texttt{\%} were positive and 50\texttt{\%} were negative reviews. Sentiment classification of Word2Vec is compared with sentiment classification of Bag-of-words. Thus, sentiment analysis can be applied in various fields.
\noindent
\\
However, this research work will employ sentiment analysis on financial news headlines related to stock market prediction or summarizing financial news headlines for stock market prediction through sentiment analysis.
\noindent
\\
There are different sentiment analysis types – Coarse-grained sentiment analysis, fine-grained sentiment analysis, emotion detection, feature/aspect-based sentiment analysis, intent analysis, and multilingual sentiment analysis. \cite{Munikar2019Fine-grainedBERT} uses a machine learning framework called Bidirectional Encoder Representations from Transformers (BERT) to explain the fine-grained sentiment classification task. \cite{Mansoor2020GlobalTime} studies binary sentiment classification (positive or negative) and emotion detection (happiness, frustration, anger, sadness, etc.). This work \cite{Karimi2020AdversarialBERT} applies adversarial training proposed by \cite{Goodfellow2015ExplainingExamples} to tweak the broad-based BERT and post-trained BERT. In this literature, a unique architecture called BAT - BERT Adversarial Training - proposed after furthering the effects of post-trained BERT with distinctive hyperparameters to apply the adversarial training for the two most essential objectives of Aspect Extraction and Aspect Sentiment Classification in sentiment analysis. The planned model surpassed both the typical BERT and post-trained BERT.

\subsection{Sentiment Analysis in Stock Market}
The Stock Market is a place where buying and selling shares of public traded companies takes place. In other words, it facilitates stock brokers to trade companies’ stocks. It is a meeting place where buyers and sellers meet. This history of the stock market dated back to 1531 in Belgium. Loan sharks and dealers would meet to deal with business, government, and personal liabilities, but in the 1500s, there was no real buying and selling of securities. In that era, many venture partnerships between businesses and financiers generated gains as the stock does. Still, there was no trading on public listed companies as there were no companies listed on the stock exchange \cite{TheExchanges}. The first company listed on the stock exchange was the Dutch East India Corporation. The first company to issue the stock and the company that established the Amsterdam Stock Exchange in 1602, now popularly known as Euronext Amsterdam \cite{WhatStock}. As the world has progressed, so do the stock markets. They deal with buying and selling shares and trade other financial securities such as bonds, derivatives, currencies, commodities, etc. Stock exchange, which is used interchangeably for the stock market, is hugely affected by political, economic, and social factors. These factors, which are also known as internal and external factors, act and move the stock market unpredictably by soaring and dropping the stock prices every second. Predictive Analytics, one of the advanced analytics subject, makes predictions about the undetermined future to solve this problem. It uses many techniques from data mining, data modeling, machine learning, AI, and deep learning algorithms to examine the existing data to make projections about the uncertain future. Predictive models capitalize on the patterns uncovered in the historical data to discover risks, prospects, and vulnerabilities. The goal is to go beyond knowing what will happen in the future and the best assessment to achieve it.

\begin{table*}
\centering
\begin{tabular}{| c | c | c | c | c | c | c | c | c |} 
 \hline
 \textbf{References} & \textbf{N-gram} & \textbf{Word2Vec} & \textbf{TF-IDF} & \textbf{TextBlob} & \textbf{BoW} & \textbf{BERT} & \textbf{Skip-gram} & \textbf{POS} \\ [0.5ex] 
 \hline\hline
 \cite{Samuels2020SentimentResponses} & $\ast$ & $\ast$ & $\ast$ & $\ast$ & $\ast$ & $\ast$ & $\ast$ & $\checkmark$ \\ 
 \cite{Rajput2020WordPandemic} & $\checkmark$ & $\ast$ & $\ast$ & $\checkmark$ & $\ast$ & $\ast$ & $\ast$ & $\ast$ \\ 
 \cite{Sahu2020SupervisedTechnique} & $\checkmark$ & $\ast$ & $\checkmark$ & $\checkmark$ & $\checkmark$ & $\ast$ & $\ast$ & $\ast$ \\  
 \cite{Sahu2016SentimentAlgorithms} & $\checkmark$ & $\ast$ & $\ast$ & $\ast$ & $\ast$ & $\ast$ & $\ast$ & $\checkmark$ \\  
 \cite{Munikar2019Fine-grainedBERT} & $\ast$ & $\ast$ & $\ast$ & $\ast$ & $\ast$ & $\checkmark$ & $\ast$ & $\ast$ \\ 
 \cite{Karimi2020AdversarialBERT} & $\ast$ & $\ast$ & $\ast$ & $\ast$ & $\ast$ & $\checkmark$ & $\ast$ & $\ast$ \\  
 \cite{Pagolu2016SentimentMovements} & $\checkmark$ & $\checkmark$ & $\ast$ & $\ast$ & $\ast$ & $\ast$ & $\ast$ & $\ast$ \\
 \cite{Medeiros2020TweetMarket} & $\ast$ & $\ast$ & $\checkmark$ & $\ast$ & $\ast$ & $\ast$ & $\ast$ & $\ast$ \\
 \cite{3Analysis} & $\checkmark$ & $\ast$ & $\ast$ & $\checkmark$ & $\ast$ & $\ast$ & $\ast$ & $\ast$ \\
 \cite{Atkins2018FinancialPrice} & $\checkmark$ & $\ast$ & $\ast$ & $\ast$ & $\ast$ & $\ast$ & $\ast$ & $\ast$ \\
 \cite{Vargas2017DeepArticles} & $\ast$ & $\checkmark$ & $\ast$ & $\ast$ & $\checkmark$ & $\ast$ & $\ast$ & $\ast$ \\
 \cite{Vargas2018DeepArticles} & $\ast$ & $\checkmark$ & $\ast$ & $\ast$ & $\checkmark$ & $\ast$ & $\ast$ & $\ast$ \\
 \cite{Dang2016ImprovementArticles} & $\ast$ & $\ast$ & $\checkmark$ & $\ast$ & $\checkmark$ & $\ast$ & $\ast$ & $\ast$ \\
 \cite{Shynkevich2015StockArticles} & $\ast$ & $\ast$ & $\checkmark$ & $\ast$ & $\checkmark$ & $\ast$ & $\ast$ & $\ast$ \\
 \cite{Akita20162016Proceedings} & $\ast$ & $\ast$ & $\ast$ & $\ast$ & $\checkmark$ & $\ast$ & $\ast$ & $\ast$ \\
 \cite{Kalyani2016StockAnalysis} & $\ast$ & $\ast$ & $\checkmark$ & $\ast$ & $\checkmark$ & $\ast$ & $\ast$ & $\ast$ \\
 \cite{Liu2018LeveragingNetwork} & $\ast$ & $\ast$ & $\ast$ & $\ast$ & $\checkmark$ & $\ast$ & $\checkmark$ & $\ast$ \\
 \cite{PDFModel} & $\ast$ & $\ast$ & $\ast$ & $\ast$ & $\ast$ & $\checkmark$ & $\ast$ & $\ast$ \\
 \cite{Fauzi2019Word2VecLanguage} & $\ast$ & $\checkmark$ & $\ast$ & $\ast$ & $\ast$ & $\ast$ & $\ast$ & $\ast$ \\
 \cite{Hajek2017CombiningReturns} & $\ast$ & $\ast$ & $\ast$ & $\ast$ & $\checkmark$ & $\ast$ & $\ast$ & $\ast$ \\
 \cite{Krauss2014PREDICTIVECASE} & $\ast$ & $\ast$ & $\ast$ & $\ast$ & $\checkmark$ & $\ast$ & $\ast$ & $\checkmark$ \\ [1ex]
\hline
\end{tabular}
\caption{Text Representation}
\label{table:ta}
\end{table*}

\subsection{Forecasting Stock Market through Emotions}
Researchers have conducted numerous intriguing studies on this particular field. Some of the analyses have shown productive results, but some haven’t displayed the most pleasing outcomes. One such study \cite{Pagolu2016SentimentMovements} concluded that a strong connection exists between stock prices of \texttt{\$}MSFT and the company’s emotions or sentiments in its tweets by using sentiment analysis and supervised machine learning principles. This study used two distinct textual concepts: Word2Vec and N-gram for understanding the public emotions in the tweets. Tweets included abbreviations, emoticons, and pointless information such as pics and URLs. For pre-processing, the researcher applied tokenization, stopwords deletion, and regex matching. The features obtained fed to the classifier and prepared using the Random Forest. However, the model trained with N-gram had an accuracy of 70.5\texttt{\%}. The model equipped with Word2Vec was chosen to categorized nonhuman annotated tweets because of its good performance on large datasets and significant meanings. 

\begin{table}[!h]
\centering
\begin{tabular}{| c | c | c |} 
\hline
\textbf{Text Representation} & \textbf{ML Algorithms} & \textbf{Accuracy} \\ [0.5ex]
\hline\hline
N-Gram & Random Forest & 70.5\texttt{\%} \\
\hline
Word2Vec & Random Forest & 70.2\texttt{\%} \\
\hline
\end{tabular}
\caption{Model Accuracy Results \cite{Goodfellow2015ExplainingExamples}}
\label{table:ta}
\end{table}
\noindent
The claim that positive sentiments in tweets about a company reflecting its stock price is strengthened by the results achieved above and confirms to have an optimistic future in further research. Similar things are being studied in other parts of the Globe. \cite{Medeiros2020TweetMarket} studies the sentiments on the Brazilian stock market respective to its tweets posted by the general public. The dataset used in this particular study contains 4516 Portuguese tweets regarding the main index of the BOVESPA. Each tweet was labeled per Plutchik’s Psych evolutionary Theory of Basic Emotions \cite{Imbir2017PsychoevolutionaryPlutchik}; however, most of the tweets appeared non-labeled, making the dataset unbalanced. The initial step sifts the tweets’ relevant words by converting them to lowercase and removing irrelevant and non-labeled tweets, known as pre-processing, and calculates TF-IDF feature vectors. Well-known methods, Principal Component Analysis (PCA) and t-Distributed Stochastic Neighbor Embedding (t-SNE), were applied to reduce TF-IDF vectors’ dimensionality and generate a visual interpretation of the tweets. K-means algorithm, Latent Dirichlet Allocation (LDA), and Non-negative matrix factorization (NMF) find insights on tweets, hence uncovering close relations and significant patterns among clusters, extricated topics, and sentiments. Since the dataset utilized is unbalanced, the process could not determine the distinguishing attributes for each emotion. However, the classification techniques were executed well in the absence of emotions. Empirical results revealed that the intended sentiment classification could foresee emotions in the tweets concerning the BOVESPA \cite{IndiceWikipedia} in Portuguese. Random Forest \cite{RandomTopics} and Support Vector Machines (SVM) \cite{SupportScience} showed the results for neutral sentiments since the number of labeled neutral tweets is higher than the number of other sentimental tweets. For SVM (neutral), precision – 61\texttt{\%}, recall – 48\texttt{\%}, F1-score 54\texttt{\%}. For Random  forest (neutral), precision – 59\texttt{\%}, recall – 40\texttt{\%}, F1-score 48\texttt{\%}. In the introduction of sentiment analysis, a study \cite{Mansoor2020GlobalTime} discussed sentiment analysis to detect people’s prevalent emotions expressing their opinions on Twitter regarding the COVID-19 pandemic. In a similar study \cite{3Analysis}, sentiment analysis on tweets related to COVID-19 forecasted market trends or predict investor reactions by understanding the emotions behind those tweets posted by users on stocks. This particular research was conducted to improve prediction capabilities and forecast recessions. This research has shown that examining internet searches to determine consumer emotions on the economy can be a precise predictor of fiscal activity and demand \cite{Vosen2011ForecastingTrends} \cite{CHOI2012PredictingTrends}. The research found out that Coronavirus’s arrival and its proliferation in a nation surged economic angst and stress—the sample set of data represented the US population. The study’s focus was on the Pharmaceutical sector since its role has been significant in the US financial industry. Researchers formed a theory that increased press coverage on the expansion of contagious disease has a positive effect on the buying and selling of pharmaceutical companies’ equity. Generally, pharmaceutical companies respond to the need for vaccines and medicines to combat contagious diseases by investing substantial amounts of funds into Research and Development. Limited appropriations awarded by governments are known for large-scale production of drugs and sanitizers, and protective masks. Data for social news and conversations were collected from Reddit, a US-based aggregator. Financial data and info were gathered from Yahoo Finance. For text processing, TextBlob was effectively suited for sentiment analysis. Its function assesses two features – polarity and subjectivity. The former ranges from -1 to 1 and later falls within 0 to 1. Using TextBlob, this research classified the news headlines as positive, negative, or neutral. After the tokenization, tagging, abbreviation processing, and n-grams - combination of words, the Naïve Bayes algorithm was applied for classification. This research paper corroborated that the model improves accuracy. Another study \cite{Atkins2018FinancialPrice} focus on the usage of bigrams, an improvement to the elementary model, that was believed to pick up a larger level of semantic info. The trigrams weren’t found to be useful because of data density problems. Obtained textual data was worked around by NLP techniques such as stemming, tokenization, etc. In other words, the research focused on using the information acquired from news to predict the stock price’s movement or future asset price’s precise value. The findings extracted from the news reports better indicated the direction of asset volatility than the price movement’s position. Two machine learning models, Latent Dirichlet Allocation (LDA) \cite{LatentWikipedia} and Naïve Bayes \cite{NaiveScience} were built to represent information from the news feeds and predicted the direction of the stock price. The results demonstrated that the forecasting accuracy for volatility was 56\texttt{\%} while that of the asset’s closing price did not exceed 49\texttt{\%} on the incoming of the latest information. Hence, the paper concluded that unpredictability measures are more likely than asset price movements. In this paper \cite{PDFModel} a BERT model is proposed on Chinese stock opinions to improve the classification of sentiments. BERT is implemented on the sentence level of the opinions. A sum of 9204 opinions or sentiments is in the dataset. To obtain a more effective classification model, different models - BERT + Linear (92.50\texttt{\%}), BERT + LSTM (91.97\texttt{\%}), and BERT + CNN (91.97\texttt{\%}) - were implemented, out of which the research chose BERT + Linear for sentiment analysis. Another research \cite{Hajek2017CombiningReturns} seeks to obtain both emotions and Bag-of-words info from the annual reports of the US-based companies. The sentiments were centered around two dictionaries, Diction 7.0 \cite{DictionSimplified} and dictionary from Loughran and McDonald \cite{ResourcesDame}. The results of Multilayer perceptron NN are compared with the effectiveness of four methods that are generally applied in text classification – naïve Bayes, decision tree, K-NN classifier, and SVM. SVM and K-NN were good on lower-dimensional data. Naïve Bayes, decision trees, and NN were better for BoW with larger features. SVM (97\texttt{\%}), Decision Tree (82\texttt{\%}), NN(101\texttt{\%}), K-NN (72\texttt{\%}), and NB (85\texttt{\%}). This paper \cite{PDFData} proposes four methodologies for forecasting stock market prices – Linear regression, SVM, Naïve Bayes, and Random Forest, out of which SVM performs the best. Tweets are collected over a particular period and might contain unnecessary information. To clear this kind of information, preprocessing such as tokenization, stopwords, regex, etc. are implemented.

\subsection{Predictive Analytics in the Stock Market}
Predictive analytics inspects, prepares, cleans, transforms data from multiple sources, and uses cutting-edge AI and Machine Learning technologies to analyze the individual stock or indices, arriving at the best conclusion to assist long-term investors in making the right decisions. Minimizing future uncertainties, making strategic decision-making, and increasing its financial competencies are critical aspects of predictive analytics. Predictive Analytics deals with remarkably diverse and extensive amounts of stock market data. It uses many algorithmic models to get the best assessment. By learning from the rich historical data, predictive analytics offers the traders or the analysts something outside standard business reports and earnings forecasts \cite{Jampala2019PredictiveSensex}. By introducing advanced technologies, traders and research analysts could obtain more accurate stock price movements. \cite{Li2020ANews} suggests a deep learning method foreseeing future stock movement. The technique combined two RNNs (LSTM and GRU), configured, and compared with the Blending Ensemble model, and the latter outpaces the former. In this paper, the stocks-related data was taken from the S and P 500 index, and news data is captured from websites such as fortune.com, cnbc.com, reuters.com, and wsj.com. From the data of the news, only the news headlines were taken into consideration as stories or articles can complement extra noises to the model, hence the model can perform poorly. The target value taken during this work was the adjusted closing price as it is deemed to be the actual price of the asset and is frequently used in a thorough evaluation of historical gains. The research shows that a blending ensemble deep learning model exceeded the best prediction model by using the same dataset. It reduced the MSE from 438.94 to 186.32, a 57.55\texttt{\%} reduction, increased F1-Score by 44.78\texttt{\%}, Recall by 50\texttt{\%}, Precision rate by 40\texttt{\%}, and MDA by 33.34\texttt{\%}. In data Pre-processing, VADER is used to calculate the sentiment scores to know the sentiment’s positivity or negatively or neutrality. The purpose of this study was to suggest that ensemble deep learning technologies can indeed predict share prices more successfully and assist traders, fund managers, and other short-term investors in making better investment decisions than other traditional methods. For future research, there are a lot of things to work around and improve the current ensemble model. Another study \cite{Vargas2017DeepArticles} proposes a DL technique that blends Convolutional neural network (CNN) with Recurrent Neural Network (RNN) for day trading directional movements projection of Standard and Poor’s 500 index using financial news articles and technical as input. This paper shows that CNN is more acceptable than RNN on spotting connotations from the content. Still, RNN is one step ahead of spotting the contextual material and building convoluted features for stock market prediction. Various models were built and compared, but EB-CNN (Event embedding input and CNN prediction model) showed better test results of 64.21\texttt{\%} relative to other models. To prepare for word embedding, the Word2Vec model is selected. A combination of BoW and SVM models is also built with an accuracy of 56.38\texttt{\%}. \cite{Vargas2018DeepArticles} have also revealed that CNN is better than RNN on capturing the lexical from content, and RNN is more refined on seeing the relative information. In this paper, two models are compared to estimate the stock price of Chevron Corporation: first is a hybrid model comprised of a Convolutional Neural Network (CNN) for the News and Long short-term memory (LSTM) for technical indicators, named as SI-RCNN, and second is an LSTM network, I-RNN, for technical indicators. These methods demonstrate a significant role of financial information in steadying the outcomes and assisting traders in deciding when to buy or sell a stock. Results indicated that the SI-RCNN could make a fair profit (13.94\texttt{\%}), and SI- RCNN has an accuracy of 56.84\texttt{\%} that is better than the accuracy of I-RNN of 52.52\texttt{\%}.
\begin{table}[!h]
\centering
\begin{tabular}{| c | c |} 
\hline
\textbf{Model} & \textbf{Test Accuracy} \\ [0.5ex]
\hline\hline
I-RNN & 52.52\texttt{\%} \\
\hline
SI-RCNN & 56.84\texttt{\%} \\
\hline
\end{tabular}
\caption{Comparison of Accuracy Results \cite{Vargas2018DeepArticles}}
\label{table:ta}
\end{table}
\noindent
A word2vec model is prepared on the bag-of-word (BoW) design. A pre-trained vector prepared on the Google News dataset was used for the word embedding technique. News articles are obtained from Reuters.com for a period between Oct 2006 to Nov 2013, and Chevron Corporation stock price series are attained from Yahoo finance. \cite{Dang2016ImprovementArticles} confirms the association between businesses news and securities. Research’s result shows that the SVM achieved up to 73\texttt{\%} accuracy in forecasting stock trends. During this research work, a trading simulation was designed to assess profitability by creating real-like conditions. The early investment was supposed 100 million VND, and each trade was levied with 0.25\texttt{\%} of the trading money for buy and sell. The system made four trades within two weeks to formulate the prediction model and earned 105,314,500 VND, a profit of over 5 million VND. For tokenization, Bag-of-words (BoW) method was used. Instead of the typical TF-IDF method, delta TF-IDF was built to improve the importance of phrases that randomly allocated positive and negative classes. Stock price-related data was collected from cophieu68.com. A total of 1884 news articles were gathered from websites such as vietstock.cn, hsx.vn, and hsn.vn. \cite{Guo2020ESG2Risk:Prediction} focused on incorporating environmental, social, and governance (ESG) events from the financial news and examining the predictive value of ESG on stock variability. A novel deep learning framework, ESG2Risk, was implemented to forecast stock market prices’ future volatility. The language-based model successfully extricated info from ESG events to predict the stock’s volatility. The model’s sentiments and test embeddings were tested for two weeks, and results showed that ESG2Risk considerably surpasses the Senti method. The investigation indicates that ESG events substantially influenced future returns and were essential for investors to consider when investing. The findings also confirmed that ESG events incorporation could build profitable investment strategies. \cite{Krauss2014PREDICTIVECASE} analyze the data aggregated from multiple online sources, including Twitter, Yahoo! Finance, and news articles. To forecast price movements, messages are filtered from stocks, sentiment analysis is applied to messages, posts, and news by applying Naïve Bayes in the combination of bag-of-words and POS tagging and accumulate sentiments for the predictor to generate trading signals for buy and sell. Finally, taking into account 833 virtual trades, the model outpaced the S and P 500 index and attained an encouraging return on investment of ~0.49\texttt{\%} per trade or ~0.24\texttt{\%} when altered by the market.

\subsection{International News in forecasting Stock Prices}
News plays an essential role in evaluating stock prices since it contains a firm’s qualitative information that influences market expectations. Uniqueness in economic news affects stock returns. There has to be something unique in the news report to move the price up or down. Both the financial figures and textual information containing novelty have a substantial effect on the stock prices. Because of the advent of the internet, the amount of data has become widely available, and many investors have found themselves overwhelmed while following the news report. Automated Document Classification of the most critical information becomes more pertinent \cite{AutomaticIs}. Automated grouping of the news report constitutes a text analysis that transforms unstructured data into a machine-readable format or in a language easily understood by the machine and then using various machine learning techniques to classify texts by emotions, topics, and objectives \cite{Arras2017WhatApproach}. A similar study, but a different experiment was conducted on news articles to predict short-term stock price movements, an old paper \cite{Gidofalvi2001UsingMovements}, but very insightful. In this paper, price movement and each news article were classified up, down, or unchanged, corresponding to the stock’s activity in a time interval surrounding the story’s publishing. A Naïve Bayesian text classifier possibly predicted the price signals of the related security. The result showed a robust connection between stock price and the news article from 20 minutes before and 20 minutes after the financial news becomes overtly attainable. Speculation among investors is commonly triggered by the announcement of a news article and outcomes in price directions. Since news articles are a compelling 
\begin{table*}
\centering
\begin{tabular}{| c | c | c | c | c | c | c | c | c | c | c |} 
 \hline
 \textbf{References} & \textbf{RF} & \textbf{SVM} & \textbf{Naïve Bayes} & \textbf{LSTM} & \textbf{LDA} & \textbf{CNN} & \textbf{DT} & \textbf{KNN} & \textbf{GRU} & \textbf{Regression} \\ [0.5ex] 
 \hline\hline
 \cite{Pagolu2016SentimentMovements} & $\checkmark$ & $\ast$ & $\ast$ & $\ast$ & $\ast$ & $\ast$ & $\ast$ & $\checkmark$ & $\ast$ & $\ast$ \\ 
 \cite{Medeiros2020TweetMarket} & $\checkmark$ & $\checkmark$ & $\ast$ & $\ast$ & $\ast$ & $\ast$ & $\ast$ & $\ast$ & $\ast$ & $\ast$ \\ 
 \cite{3Analysis} & $\ast$ & $\ast$ & $\checkmark$ & $\ast$ & $\ast$ & $\ast$ & $\ast$ & $\ast$ & $\ast$ & $\ast$ \\  
 \cite{Atkins2018FinancialPrice} & $\ast$ & $\ast$ & $\checkmark$ & $\ast$ & $\checkmark$ & $\ast$ & $\ast$ & $\ast$ & $\ast$ & $\ast$ \\  
 \cite{Li2020ANews} & $\ast$ & $\ast$ & $\ast$ & $\ast$ & $\checkmark$ & $\ast$ & $\checkmark$ & $\ast$ & $\ast$ & $\ast$ \\ 
 \cite{Vargas2017DeepArticles} & $\ast$ & $\ast$ & $\ast$ & $\checkmark$ & $\ast$ & $\checkmark$ & $\ast$ & $\ast$ & $\ast$ & $\ast$ \\  
 \cite{Vargas2018DeepArticles} & $\ast$ & $\ast$ & $\ast$ & $\checkmark$ & $\ast$ & $\checkmark$ & $\ast$ & $\ast$ & $\ast$ & $\ast$\\
 \cite{Dang2016ImprovementArticles} & $\ast$ & $\checkmark$ & $\ast$ & $\ast$ & $\ast$ & $\ast$ & $\ast$ & $\ast$ & $\ast$ & $\ast$\\
 \cite{Shynkevich2015StockArticles} & $\ast$ & $\checkmark$ & $\ast$ & $\ast$ & $\ast$ & $\checkmark$ & $\ast$ & $\ast$ & $\ast$ & $\ast$ \\
 \cite{Akita2016DeepInformation} & $\ast$ & $\ast$ & $\ast$ & $\checkmark$ & $\ast$ & $\ast$ & $\ast$ & $\ast$ & $\ast$ & $\ast$ \\
 \cite{Kalyani2016StockAnalysis} & $\checkmark$ & $\checkmark$ & $\checkmark$ & $\ast$ & $\ast$ & $\ast$ & $\ast$ & $\ast$ & $\ast$ & $\ast$ \\
 \cite{Liu2018LeveragingNetwork} & $\ast$ & $\checkmark$ & $\ast$ & $\checkmark$ & $\ast$ & $\checkmark$ & $\ast$ & $\ast$ & $\ast$ & $\ast$\\
 \cite{PDFModel} & $\ast$ & $\ast$ & $\ast$ & $\checkmark$ & $\ast$ & $\checkmark$ & $\ast$ & $\ast$ & $\ast$ & $\ast$ \\
 \cite{Fauzi2019Word2VecLanguage} & $\ast$ & $\checkmark$ & $\ast$ & $\ast$ & $\ast$ & $\ast$ & $\ast$ & $\ast$ & $\ast$ & $\ast$ \\
 \cite{Hajek2017CombiningReturns} & $\ast$ & $\checkmark$ & $\checkmark$ & $\ast$ & $\ast$ & $\checkmark$ & $\checkmark$ & $\checkmark$ & $\ast$ & $\ast$ \\
 \cite{Krauss2014PREDICTIVECASE} & $\ast$ & $\ast$ & $\checkmark$ & $\ast$ & $\ast$ & $\ast$ & $\ast$ & $\ast$ & $\ast$ & $\ast$ \\
 \cite{Aamir2017StoryMarket} & $\ast$ & $\ast$ & $\ast$ & $\ast$ & $\ast$ & $\ast$ & $\ast$ & $\ast$ & $\ast$ & $\ast$ \\
 \cite{Kollintza-Kyriakoulia2018MeasuringTechniques} & $\ast$ & $\checkmark$ & $\ast$ & $\ast$ & $\ast$ & $\ast$ & $\ast$ & $\ast$ & $\ast$ & $\checkmark$ \\
 \cite{PDFData} & $\checkmark$ & $\checkmark$ & $\checkmark$ & $\ast$ & $\ast$ & $\ast$ & $\ast$ & $\ast$ & $\ast$ & $\checkmark$\\ [1ex]
\hline
\end{tabular}
\caption{Comparison between different machine learning techniques in the review papers}
\label{table:ta}
\end{table*}
\noindent
source of information in prediction, they influence the market by increasing and dropping prices. Researchers have been developing price prediction models established on the info disengaged from the news articles.
This research \cite{Shynkevich2015StockArticles} employed multiple kernel learning (MKL) methodology to merge information from a particular stock and sub-industry-specific news reports to project imminent price movement. For evaluation purposes, SVM with various kinds of kernels and KNN to Stock-specific datasets and sub-industry-specific datasets were applied. For feature extraction, the Bag-of-Words technique was used. During this work, the transformation of each document into a vector was characterized by TF-IDF. \cite{Akita2016DeepInformation} proposed a strategy that converted newspaper coverage into Paragraph Vector \cite{Le2014DistributedDocuments} and modeled previous incidents’ temporal effects on several companies’ opening prices with LSTM. The study demonstrated the data of 50 companies listed on TYO/TSE (Tokyo Stock Exchange), predicting the asset prices by utilizing distributed representations of news reports. Empirical results confirmed that distributed models of word-based material were better than the mathematical-data-only approach and Bag-of-words methodology. LSTM successfully captured the time-series effect of input data and can take the same-industry companies for stock price prediction. The news dataset was taken from Nikkei newspaper from 2001 to 2008, and the dataset of the top 10 companies was taken from Nikkei 225 linked with the news articles about them in the same period. This paper \cite{Kalyani2016StockAnalysis} attempted to study the correlation between news and stock movement. For detection of emotion and representation of text, both BoW and TF-IDF were used, and three different classification models – RF, SVM, and Naïve Bayes – explained the polarity’s positive or negative texts. The research concluded that stock trends could be forecasted using financial news articles and past historical stock prices. \cite{Liu2018LeveragingNetwork} worked on the sentiment signal features by utilizing deep neural models to extract rich lexical characters from the news text. To encode text and to get the contextual information, a Bidirectional-LSTM was applied. The At-LSTM model predicted the S and P 500  index’s directional movements and different companies’ stock prices used financial news titles. During this work, the maximum accuracy for this planned model was 65.53\texttt{\%}, and the average accuracy was 63.06\texttt{\%} lower than the KGEB-CNN (Knowledge graph event embedding (KGEB) \cite{Ding2016Knowledge-DrivenPrediction}. Empirical results suggested that the model is beneficial and viable with the state-of-the-art model. Future work proposed looking at predicting price movements at a different time-frame and relates to the headlines. For word and phrase embedding, skip-gram and Bag-of-words were applied. \cite{Aamir2017StoryMarket} investigated the relationship between news reports and stocks. It attempted to understand whether news connects with the KSE-100 index. This study employed two methods – correlation and regression analysis – to a word list and KSE. For this study, 16 years of data were used (1999-2014). The result was that words had a casual relationship with both the KSE and the index. \cite{Kollintza-Kyriakoulia2018MeasuringTechniques} analyzed and modeled the effects of technical analysis, online news, and tweets on predicting the stock’s value. This paper studied the relationship between the time series of the closing price and news reports, and the time series of the closing price and opinions on Twitter. Two methods – linear regression and SVM regression – were implemented, but SVM achieved better results than linear regression. 

\begin{table*}[!h]
\centering
\begin{tabular}{| c | c | c |} 
 \hline
 \textbf{References} & \textbf{Data Source} & \textbf{Dataset Details} \\ [0.5ex] 
 \hline\hline
 \cite{Pagolu2016SentimentMovements} & \makecell{Twitter API \\ Yahoo Finance} & \makecell{- Period: 31st Aug 2015 to 25th Aug 2016. \\ A total of 2,50,000 tweets on Microsoft. \\ - Stock opening and closing prices of \texttt{\$}MSFT \\ between the same period from Yahoo \\ Finance extracted.} \\ 
 \hline
  \cite{Medeiros2020TweetMarket} & \makecell{Public Domain} & \makecell{The dataset contains 4516 Portuguese tweets \\ regarding the main index of the BOVESPA.} \\ 
   \hline
  \cite{3Analysis} & \makecell{Reddit \\ Yahoo Finance} & \makecell{- Data for social news and conversations \\ are collected from Reddit. - Financial data \\ and news gathered from Yahoo Finance} \\  
   \hline
  \cite{Atkins2018FinancialPrice} & \makecell{Bannot Gang \\ Yahoo Finance \\ Reuters} & \makecell{- min-by-min intraday data downloaded from \\ the quant trading website ‘The Bannot Gang.’ \\ - Missing daily data imputed from Yahoo \\ finance database. - Textual data from Reuters US} \\ 
  \hline
  \cite{Li2020ANews} & \makecell{CNBC \\ Reuters \\ WSJ \\ Fortune} & \makecell{ - The news data is captured from the mentioned websites.} \\ 
  \hline
  \cite{Vargas2017DeepArticles} & \makecell{Reuters \\ Yahoo Finance} & \makecell{ - Obtained 106494 news articles from the Reuters website. \\ - S and P 500 index was selected from Yahoo finance.} \\ 
  \hline
  \cite{Vargas2018DeepArticles} & \makecell{Reuters \\ Yahoo Finance \\ Google.com Archive \\ Word2Vec } & \makecell{ -The Chevron corporation stock price data was \\ obtained from Yahoo Finance for the same phase. \\ - Google News data (about 100 billion words)} \\
  \hline
  \cite{Dang2016ImprovementArticles} & \makecell{LexisNexis database  \\ Business Wire \\ PR Newswire \\ McClatchy- Tribune \\ Business News \\ Yahoo Finance } & \makecell{ - Data collected from 1st Sept 2009 to 1st Sept 2014 \\ - About 8264 financial news articles were downloaded \\ from LexisNexis Database. - Business Wire, PR \\ Newswire, and McClatchy-TBN were selected as \\ providers of news stories because they showed \\ sufficient press coverage of stocks constituting \\ the S and P 500 market index. - Data of historical \\ prices was downloaded from Yahoo Finance} \\ 
  \hline
  \cite{Shynkevich2015StockArticles} & \makecell{Nikkei Newspaper \\ Nikkei 225} & \makecell{ - Morning edition of Nikkei newspaper from 2001 to 2008 \\ - 10 companies were selected from Nikkei 225. Companies \\ appearing often in news from 2001 to 2008 } \\
 \hline
 \cite{Akita2016DeepInformation} & \makecell{New.google.com \\ Reuters.com \\ Yahoo Finance} & \makecell{- collected Apple Inc. data from 1st Feb 2013 to 2nd \\ Apr 2016. - news stories related to Apple inc and daily \\ stock 
prices of AAPL for the same period.} \\ 
\hline
 \cite{Kalyani2016StockAnalysis} & \makecell{Bloomberg \\ Reuters \\ Yahoo Finance} & \makecell{- News articles were obtained from Reuters and \\ Bloomberg from Oct 2006 to Nov 2013. - 473 companies' \\ data listed on S and P 500 were taken from Reuters. \\ - Companies stock price information from Yahoo} \\ 
 \hline
 \cite{Aamir2017StoryMarket} & \makecell{Pak and \\ Gulf Economist} & \makecell{- financial and business articles \\ of the last 16 years from the magazine.} \\ 
  \hline
 \cite{PDFData} & \makecell{KSE \\ Twitter} & \makecell{-KSE 100 index - Tweets collected from Twitter} \\[1ex]
\hline
\end{tabular}
\caption{Dataset used in various Review Papers}
\label{table:ta}
\end{table*}
\noindent
 
\begin{table*}[!h]
\centering
\begin{tabular}{| c | c |} 
\hline
\textbf{Question no} & \textbf{Research Questions} \\ [0.5ex]
\hline\hline
Q1 & \makecell{Comparison between different sectors (such as Oil and Gas, Banking, IT, etc.) of the economy concerning \\ the financial news and which one performs better?} \\
\hline
Q2 & \makecell{Which company is the best and the worst performer in its sector concerning the financial news?} \\
\hline
Q3 & \makecell{How do positive sentiments lead to buying, negative emotions lead to selling and neutral leads to nothing?} \\ 
\hline
Q4 & \makecell{Are these indexes related to each other, i.e., does one affects the other or not?} \\ 
 \hline
Q5 & \makecell{Is there a way out when global news does not affect the local indices such as India (nifty 100) \\ or does the news around the Globe impact all indices?} \\ 
\hline
Q6 & \makecell{Which index such as nifty100, S and P 500, FTSE100, etc. performs better with the Global Financial \\ News, helping investors decide where to invest?} \\ 
\hline
Q7 & \makecell{Which Deep Learning model - RNN, or LSTM, performs better on sentiment analysis?} \\
\hline
Q8 & \makecell{Comparing word embedding methods – Word2Vec, FastText, Glove, BERT, TF-IDF, Bag of Words, and \\ which one is the best?} \\ [1ex]
\hline
\end{tabular}
\caption{Research Questions}
\label{table:ta}
\end{table*}
\noindent
\section{Research Questions}
The main intent of this research is to find answers to some questions that might have been explored by many researchers, analysts, fund managers, investment advisors, and retail investors. Throughout the inception of the stock market, almost all the investors are looking to make as much profit as possible, but some professional and experienced researchers are looking to solve some real-world problems such as how to predict the correct equity returns? How to predict the next stock market crash? When is the next boom and bust? How inflation affects the economy? Which country’s economy plays a key role in the global economy? How rational is it to play in the market? These are few questions that have been intriguing to many researchers and numerous questions have already been answered. Getting answers to these questions assists in improving the imagination that came through curiosity. Here in this research, we are looking for the answers to some detailed-oriented questions that might have been already answered by a few researchers, but we are looking for some improved solutions or whether the answers already known has proof of being valid. This research has the curiosity to know about different sectors of the economy related to the news, best and worst-performing companies, indices relationship with each other, out of top indices, which one performs better?, best performing deep learning model and word embedding methods. In this research work, these are the key questions as given in Table 6.

\section{Overview}
This paper aims to understand how major stock market indices react to the emotions/sentiments extracted from the financial news headlines. This objective makes this research categorized into "Explanatory Research,” also called “Casual Research,” because of the cause and effect relationship between news sentiments and stock market indices. This research can also be characterized as “Correlational Research” and “Descriptive Research” to understand if any relationship between different indices exists and how positive sentiments lead to buying, negative emotions lead to selling, and neutral leads to nothing. This research consists of four main sections as shown in Fig. 7, first is the Data Preprocessing techniques that are used to convert the raw data into a useful and effective format. The second section is VADER, a lexical-based approach to calculate sentiment scores. A technique that assists in calculating emotions in streaming media such as text, audio, or video. The other section is the NLP techniques where words or phrases from the dictionary are mapped to real numbers vectors. The techniques used in this work – BoW, TF-IDF, Word2Vec, BERT, GloVe, and FastText. The final section is building deep learning models – RNN and LSTM – to analyze those sentiments.

\section{Data Preprocessing}
The following are the approaches that are used in the process of data cleaning and preprocessing.

\subsection{Lowercasing}
An essential step in data cleaning is to convert the text into lowercase. Lowering the case is extremely helpful in reducing feature vectors.

\subsection{Stemming}
It is the process of reducing derived words to their word stem, base, or root form. For example, A stemming algorithm may reduce the words dogs, doglike, and dogy to the stem dog \cite{StemmingLemmatization}.

\subsection{Lemmatization}
It is the method of determining the lemma of a word based on its proposed meaning. Lemmatization relies on identifying the correct part of speech and the definition of a word in a sentence and its surrounding sentences. For example, the verb “to play” may appear as “play,” “played,” “plays,” or “playing.” The base form, “play,” is called the lemma for the word \cite{StemmingLemmatization}.

\subsection{Stopwords Removal}
Stopwords is a process of filtering out words before or after processing language text to help index and parse the text faster. Example: “a,” “an,” “the,” “but,” “what.” \cite{DroppingWords}

\subsection{Normalization}
It is the process of converting a text into a canonical form. For example, 2moro, 2mrrw, 2morrow, “tomrw” transformed into “tomorrow.” 

\subsection{Tokenization}
Tokenization splits a piece of text into smaller chunks such as words or sentences. It is called tokens. Tokenization is generally categorized into three types - word, character, and sub-word (n-gram characters) tokenization. For example, consider this sentence: “Give your best shot.” The way to form tokens is dependent on space acting as a delimiter. The tokenization leads to four tokens – Give-your-best-shot. Since each token is a word, it becomes an illustration of Word tokenization \cite{TokenizationKeras}.
\begin{figure}[h]
\centering
  \includegraphics[width=.5\textwidth]{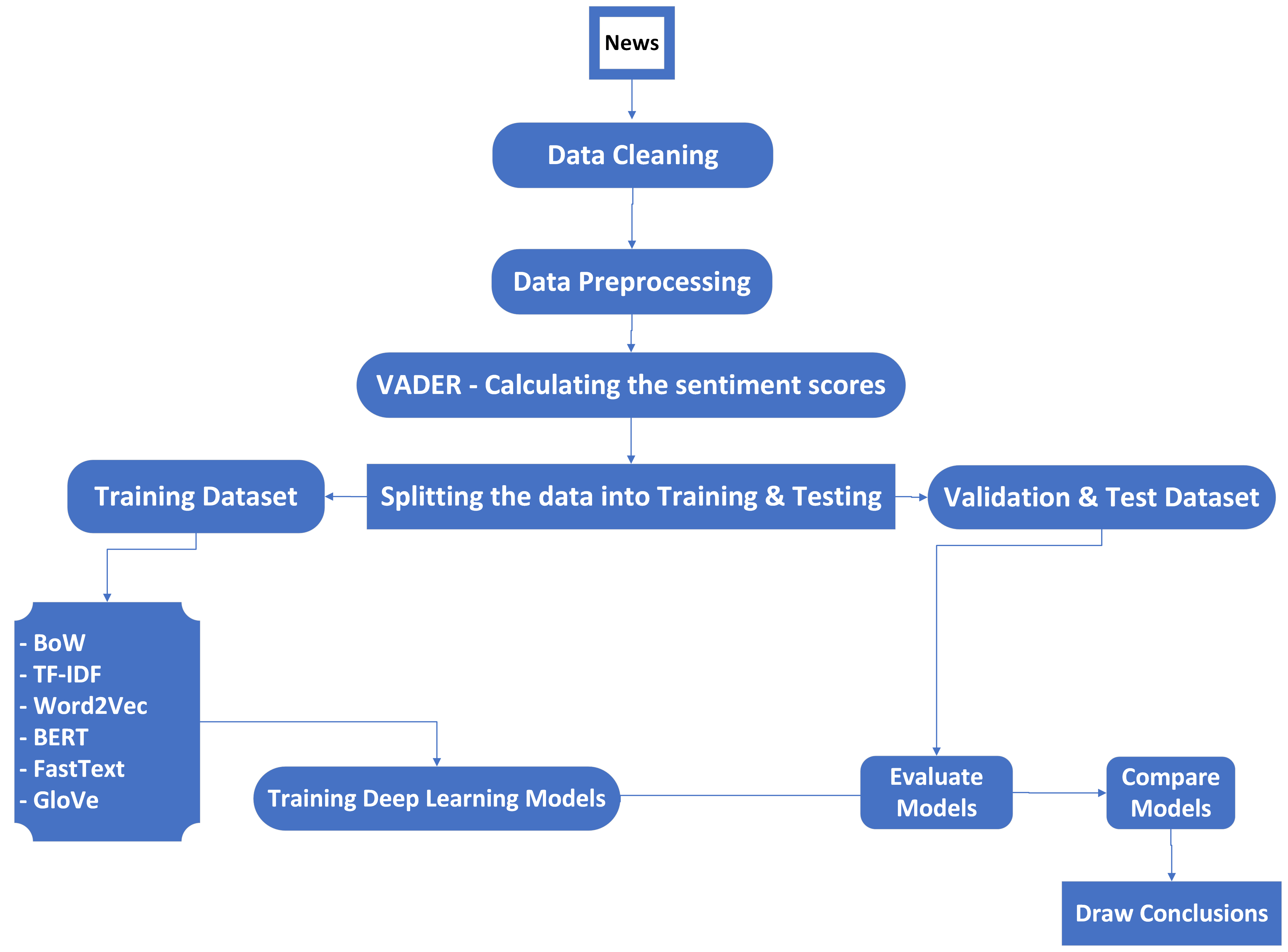}
\caption{Overview or Flow of Work}
\label{fig:exampleFig}
\end{figure}
\subsection{Part of Speech (POS) Tagging}
POS Tagging seeks for connections within the sentence and gives a corresponding tag to the word. This undertaking is not effortless since a specific phrase may have a separate  part of speech centered on the context. For example, In the sentence “Give me your answer,” the answer is a noun, but in the sentence “Answer the question,” the answer is a verb \cite{NLPMedium}.

\subsection{Special Character Removal}
The advent of email, social media, and text messaging have given rise to text-based emoticons represented by ASCII special characters. For example, =) or $>$:( is very indicative of sentiment because they directly reference happiness or sadness. Stripping our messages of these emoticons by removing special characters will also strip meaning from our message. Usually, you might be dealing with accented characters/letters in any text corpus, especially if you only want to analyze the English language. Hence, we need to make sure that these characters are converted and standardized into ASCII characters. A simple example, convert é to e. (white spaces) \cite{NLP:Science}. Missing and Duplicate Values - It is one of the cleaning data features, a necessary step before exploratory data analysis and model building.

\section{VADER-Valence Aware Dictionary for Sentiment Reasoning}
VADER, a popular lexical-based approach, is used to calculate sentiment scores. The term ‘lexical,’ a Greek word, is a synonym for ‘expression.’ In this approach, a dictionary of lexicons is prepared. Each entry in the dictionary is assigned a sentiment score. For example, the word “good” has a positive sentiment score, and the word “bad” has a negative sentiment score. VADER, a python package, is an efficient technique that helps decipher and compute the emotions in streaming media such as text, audio, or video. Few studies have revealed 99 \texttt{\%} precision in sentiment score calculation \cite{VADERTrading}.

\begin{figure}[h]
\centering
  \includegraphics[width=.2\textwidth]{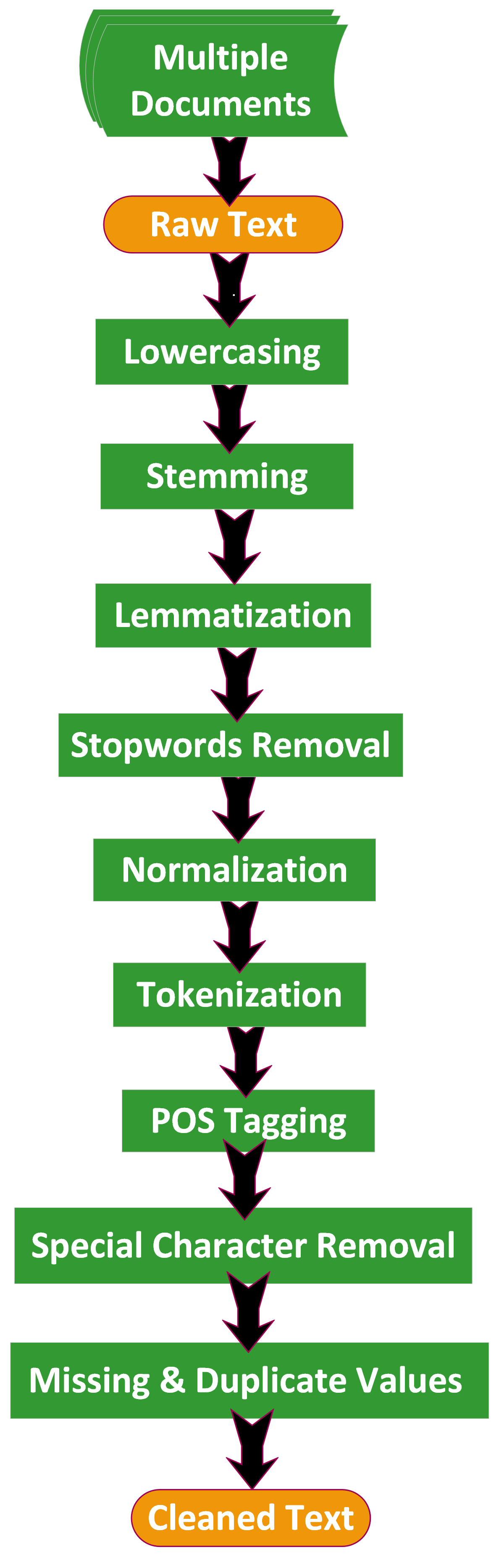}
\caption{Data Preprocessing – Text Cleaning Pipeline}
\label{fig:exampleFig}
\end{figure}

\section{Word Embeddings}
Word embedding is an NLP technique where words or phrases from the dictionary are mapped to real number vectors. It’s been noted that terms and phrase embeddings boost NLP performance, such as sentiment analysis and syntactic parsing \cite{IntroductionScience}. \\
The commonly used technique which converts words into their embedded form:

\subsection{BoW (Bag-of-Words)}
BoW, an algorithm, counts how many times a word appears in a document. These word counts allow one to compare records and gauge their similarities for applications such as search and document classification \cite{Kasthuriarachchy2014EnhancedAnalysis}.

\subsection{TF-IDF}
TF-IDF stands for Term Frequency-Inverse  Document Frequency. The BoW is a perfectly acceptable model to convert raw text to numbers. However, if the purpose is to identify signature words in a document, there is a better transformation. The TF-IDF measures relevance, not frequency. The TF-IDF scores replace word counts across the whole dataset \cite{Qaiser2018TextDocuments}.

\subsection{Word2Vec}
Word2Vec is a popular method to generate word embeddings. Word embedding converts the words into a vector space. With these vectors, we can do multiple operations like addition, subtraction, multiplication, etc. Word2Vec takes a  large corpus of text as its input and produces a vector space, typically of several hundred dimensions. Each unique word in the collection is assigned a corresponding vector in the space. In word2vec, word vectors are positioned in the vector space, terms that share  familiar contexts in the corpus are located close to one another. Word2vec can predict a word’s meaning based on past appearances. For example, the vector King minus the vector Man plus the vector Woman gives the word Queen's vector \cite{IntroductionScience}.
\noindent
\begin{figure}[h]
\centering
  \includegraphics[width=.55\textwidth, height=6cm]{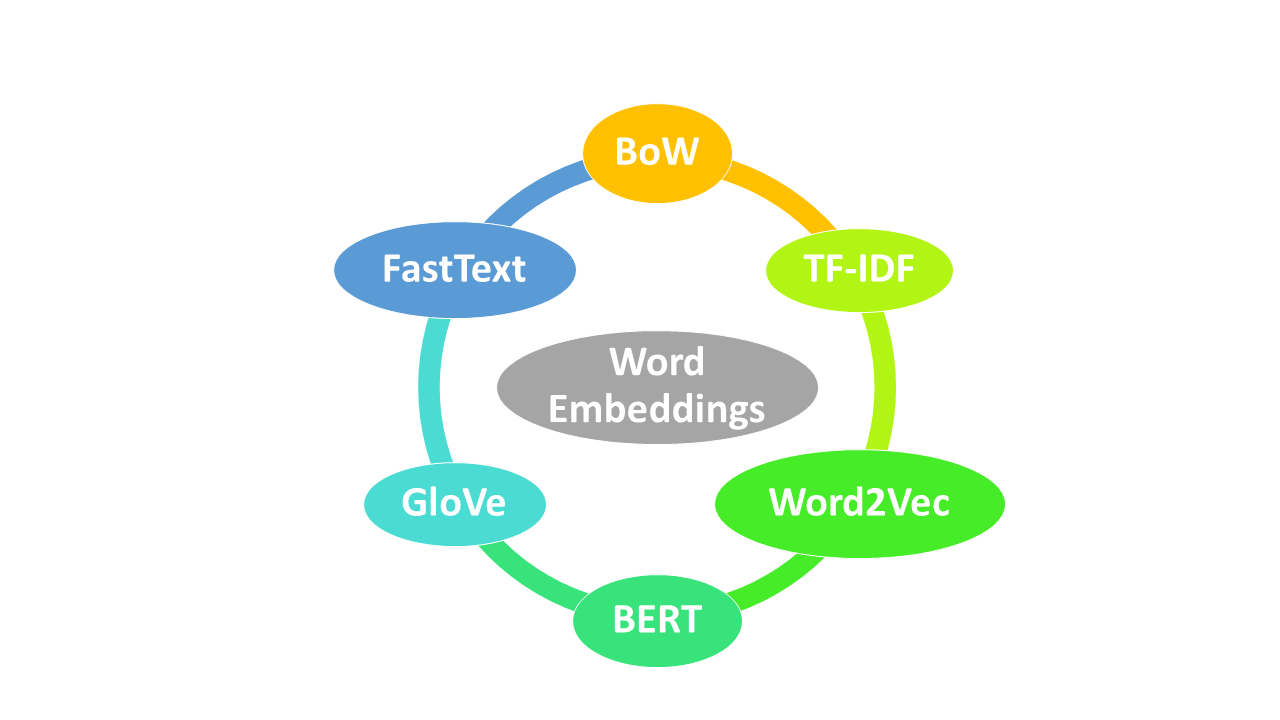} 
\caption{Different Word Embedding Techniques}
\label{fig:exampleFig}
\end{figure}
\subsection{BERT}
BERT stands for Bidirectional Encoder Representation from Transformers. The model was published in 2018 by researchers of Google \cite{GoogleProcessing}. The BERT makes use of a Transformer, an attention mechanism that learns contextual relations between words or sub-words in a text. As opposed to directional models, which reads the text input sequentially either left-to-right or right-to-left, the transformer encoder reads the entire sequence of words at once. Although it is considered bi-directional, it would be more accurate to say that it is non-directional, allowing the BERT model to learn the word context in all of its surroundings \cite{Devlin2018BERT:Understanding}.

\subsection{Glove}
Global Vectors for Word Representation is an unsupervised learning algorithm for attaining vector representation for words. Training is performed on aggregated global word-to-word co-occurrence statistics from a corpus, and the resulting illustrations showcase interesting linear substructures of the word vector space \cite{Pennington2014GloVe:Representation}.

\subsection{FastText}
FastText is a collection for studying word embeddings and content categorization. Facebook originally developed it to build unsupervised or supervised learning models to attain vector depiction for phrases \cite{Kasthuriarachchy2014EnhancedAnalysis}.

\section{Model Building}
The next section is the model building that includes the following subsections involved in creating algorithms and their evaluation.

\subsection{Neural Network}
A neural network typically takes input from all the features and figures out each neuron’s function to improve prediction accuracy. This paper focuses on Neural Network Techniques.
\\
\noindent\\
Deep learning has created a deep neural network to predict future price trading instruments using the historical OHLCV data. A deep neural network typically works with more layers than an artificial neural network. This depth helps in better learning of the model and may result in more accurate predictions. For instance, when analyzing historical price data, the first layer’s neurons might learn to recognize the trends. The neurons in the second layer could learn to recognize complex features such as Open and Close prices. The third layer may identify more complex features, and so on. This model uses three kinds of layers: dense layer, dropout layer, and activation layer \cite{Thakkar2021ADirections}.
\\
\noindent\\
Compared to Deep Neural Networks, RNNs solve a different set of real-world problems, such as Time Series prediction \cite{Petnehazi2018RecurrentForecasting}. In a time series, the data points are independent. For example, a neural network is trained to recognize images of cats. During this process, the order in which various training data images are passed into the model would not matter. It is because image-1 and image-2 are independent of each other. However, if the problem is to predict future returns or prices using financial data, the model has to understand its underlying trend. The share prices for day one and day two are not entirely independent, and this relationship needs to be captured by the model. An RNN does precisely that \cite{Pawar2019StockRNN}. An RNN can use the information from the past data by saving it in memory. In an RNN neuron, there are two points. One is the memory, and the other is the current data. To increase the memory, if one tries to increase the number of neurons in an RNN, it will lead to the exploding of gradient descent. One of the RNN model's significant drawbacks \cite{PascanuOnNetworks}. This drawback was fixed later by creating LSTMs or Long Short Term Memory models. LSTMs solve the exploding and vanishing gradient problem faced by RNNs \cite{OVERCOMINGNETWORKS}. The opposite of a vanishing gradient is called an exploding gradient. It happens when you have an activation function whose gradient or derivative is more than 1; then, your compounded final value will be immense. It makes the model very sensitive to small changes in the data and causes the algorithm to overfit. The vanishing and exploding gradients can be considered the underfitting and overfitting problems faced in most machine learning algorithms.

\begin{table*}[!h]
\centering
\begin{tabular}{| c | c |} 
\hline
\textbf{Data Source} & \textbf{Dataset Details} \\ [0.5ex]
\hline\hline
\cite{BusinessHeadlines} & \makecell{Indian Financial News articles (2003-2020) taken from the Business-Standard website.} \\
\hline
\cite{FindKaggle} & \makecell{The dataset from Kaggle has fetched news from the Business-Standard website \\ and S and P 500 companies (2013-2018)} \\
\hline
\cite{EconomicData.world} & \makecell{Financial news article tone. The dataset has tone’s judgement on a scale of 1 to 9.} \\ 
\hline
\cite{YahooNews} & \makecell{Nifty 100 companies (2000-2020);
S and P 500 companies (2000-2020)} \\ 
 \hline
\cite{NSELtd.} & \makecell{Information on the Nifty 100 companies such as symbol, company name, industry, etc.} \\ 
\hline
\cite{S-and-p-500-companies-financials/dataGitHub} & \makecell{Information on the S and P 500 companies such as symbol, name, and the sector.} \\ 
\hline
\cite{FTSEFinance} & \makecell{UK’s FTSE 100 companies (2000-2020)} \\
\hline
\cite{EURONEXTFinance} & \makecell{Europe’s EURONEXT100 companies (2000-2020)} \\
\hline
\cite{SSEFinance} & \makecell{Shanghai’s SSE50 companies (2000-2020)} \\ 
\hline
\cite{DAXFinance} & \makecell{Germany’sDAX30 companies (2000-2020)} \\ 
\hline
\cite{NikkeiFinance} & \makecell{Tokyo’s Nikkei225 companies (2000-2020)} \\ 
\hline
\cite{GoogleNews-vectors-negative300.bin.gzDrive} & \makecell{Word2Vec model trained on Google news data (about 100 billion words).} \\ [1ex]
\hline
\end{tabular}
\caption{Dataset used in this Research Work}
\label{table:ta}
\end{table*}
\noindent
\section{Dataset Description}
The data used for this particular study is retrieved from various sources such as Business Standard \cite{BusinessHeadlines}, Kaggle Dataset \cite{FindKaggle}, Data.World \cite{EconomicData.world}, Yahoo Finance \cite{YahooNews}, NSE India \cite{NSELtd.}, Github \cite{S-and-p-500-companies-financials/dataGitHub}, and google’s word2vec \cite{GoogleNews-vectors-negative300.bin.gzDrive}. Global Stock market indices consist of seven attributes – Date, open, high, low, close, volume, adjusted volume/adjusted close, and name. For some companies, the data is from the year 2000, but for some other companies, the information is from the year they got listed in the stock market. Kaggle is used to obtain companies’ historical datasets, but they are available for only 5 years. So, this study had two options either to purchase the historical dataset of 20 years, which is very expensive \cite{HomepageQuantPedia}, or to get historical data from Yahoo Finance \cite{YahooNews}; others are the list of components collected from various other sources. Google’s word2vec model \cite{GoogleNews-vectors-negative300.bin.gzDrive} trained on Google news data (about 100 billion words) to embed words or phrases while doing sentiment analysis on the stock market. Training own word vectors would be better, but it will take a long time and require more fast resources. This pre-trained Google word2vec model contains 3 million words and phrases and fits in using 300-dimensional word vectors.
\noindent\\
Global Stock market indices consist of seven attributes – Date, open, high, low, close, volume, adjusted volume/adjusted close, and Name.
\begin{table}[!h]
\centering
\begin{tabular}{| c | c |} 
\hline
\textbf{Attributes} & \textbf{Details} \\ [0.5ex]
\hline\hline
Date & Date - Stock price going up or down \\
\hline
Open & The price at which a stock traded \\
\hline
High & The highest price \\ 
\hline
Low & The lowest price \\ 
 \hline
Close & The Closing price \\ 
\hline
Volume & The total shares traded \\ 
\hline
Adj.Close & adjusted closing price \\
\hline
Name & A ticker symbol or stock symbol \\  [1ex]
\hline
\end{tabular}
\caption{Global Stock Market Indices Attributes}
\label{table:ta}
\end{table}
\noindent

\section{Dataset Preparation}
During the data preparation stage, numerous techniques were applied to the datasets. The data's lucidity was established at each phase to confirm that no info gets missed during these stages or phases. The analysis of the data preparation is discussed in various sections below:

\subsection{Elimination of Variables}
An “unnamed” column in the Indian Financial News dataset is dropped since it does not add any value for the analysis and model building. Another variable, “Description,” from the same dataset is not included in the analysis since the entire research depends on the news title or headlines. So, the two features, “Date” and “Title,” that assist in algorithm building are extracted from the Indian Financial News.
A similar dataset - Full Economic News - fetches news details from WSJ (Wall Street Journal)  and WAPO (The Washington Post), including various features; however, only a few of them add value to the model’s building. 
Features included in the model building from this dataset:
“Date,” “article id,” and “headline.” Some variables were thought to add value but later found out to have null values, so they were dropped from the dataset. “positivity” and “positivity: confidence” have 82.25\texttt{\%} and 52.81\texttt{\%} null values, so they were also dropped. Another reason to eliminate these kinds of variables, such as “relevance” and “relevance: confidence,” is that there was no more information on these labels and wasn’t helping in using VADER to calculate “Negative,” “Neutral,” “Positive,” and to calculate “Compound” scores to get buy and sell decisions. After fetching important columns from Indian Financial News and Full Economic News and assigning them to Global headlines and Indian title data frames, these new data frames have 8000 and 50,000 entries.

\subsection{Transformation of Variables}
Most of the distinct columns are validated, and no changes are needed. No inconsistency is found in these label names and has unique values used in the dataset. Still, some dataset features required transformation before being included for analysis and model building.

\subsection{Identification of Missing Values}
Although there were some missing values in the columns of the “Indian titles” and “Global Headlines” data frame, they did not assist in the model building, so they were dropped. For the rest of the implementation, no significant missing values were there for treatment.

\subsection{Derived Variables}
Two datasets used in this research proposal, “IndianFinancialNews1” and “Full- Economic-News,” consisted of 4 columns (IndianFinancialNews1) and 15 columns (Full-Economic- News). Each of these datasets was converted into a data frame, “Indian titles” and “Global Headlines,” extracting columns Date and Title for Indian titles and Date and headline for Global headlines. The rest of the 12 columns didn’t help in building the model, so they were dropped. 

\section{Exploratory Data Analysis and Data Visualization}
Exploratory Data Analysis, also known EDA, is debatably the most crucial phase in data analysis. In this research, data is dug to withdraw valuable information from it, and inferences are made \cite{WhatScience}.

\subsection{Univariate Analysis}
“Uni” refers to “one.” It is an analyzing technique that evaluates only one variable, so there is no question of causes or relationships. The entire focus of univariate analysis is one variable \cite{Canova2017HowData}. Here in this study, Tech Stocks were segregated from “Global Headlines” and Banking Stocks from “Indian Titles. The tech stocks considered were FAANG (Facebook, Amazon, Apple, Netflix, Google) and equities of other major promising companies such as Microsoft, Cisco, and Oracle. Fig. 10 and Fig. 11 shows the pie chart of tech companies with both the billion-dollar and trillion-dollar capitalization \cite{FAANGDefinition}.\\\\
Tesla has 834 billion of dollars capitalization present, as recorded at the start of the year 2021. The company’s inclusion in the S and P 500 has also given the stock a considerable edge to be one of the best-performing equities \cite{NowMarketWatch}.
\\\\
As shown in Fig. 11, Apple has a maximum trillion-dollar capitalization of 2 trillion dollars as recorded at the start of 2021. Microsoft is second with 1.66 trillion dollars in market capitalization. This analysis clearly shows that the tech stocks are just rocking.
HDFC Bank has the largest capitalization of Rs. 807,615.27 crores as shown in Fig. 12, which signifies the most prominent company in the banking sector. It is the largest private sector bank by assets. It has launched various digital products, such as Payzapp and SmartBUY.
\begin{figure}[h]
\centering
  \includegraphics[width=.4\textwidth]{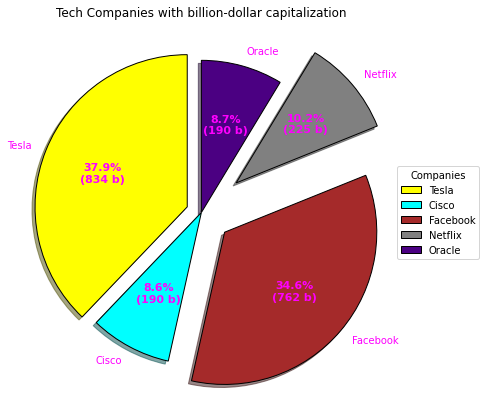}
\caption{Tech Companies with billion-dollar capitalization}
\label{fig:exampleFig}
\end{figure}

\begin{figure}[h]
\centering
  \includegraphics[width=.4\textwidth]{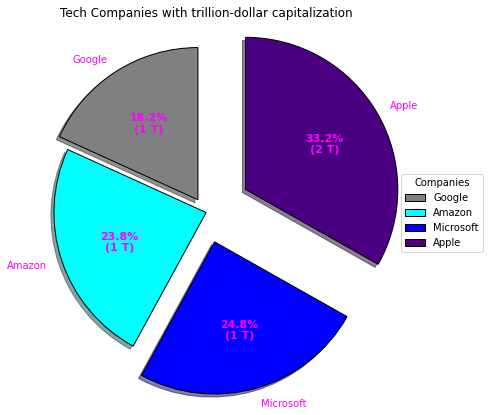}
\caption{Tech Companies with trillion-dollar capitalization}
\label{fig:exampleFig}
\end{figure}

\subsection{Bivariate Analysis}
Bivariate analysis is a quantitative statistical analysis that involves evaluating two variables to determine the experiential relation between them \cite{Sandilands2014BivariateAnalysis}. Here in this study, after calculating the compound score for each news headline or title via VADER, the suggested action taken for the negative compound score is sell, and taken for the positive compound score is buy. Fig. 13 and fig. 14 exemplify Facebook data taken from the "Global Headlines.”

\begin{figure}[h]
\centering
  \includegraphics[width=.4\textwidth]{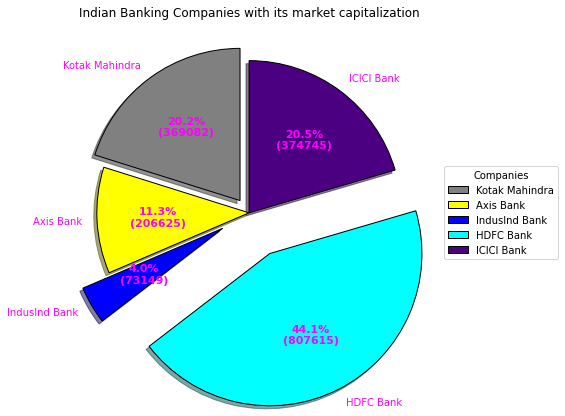}
\caption{Indian Banking Companies with its market capitalization}
\label{fig:exampleFig}
\end{figure}
\noindent
Fig. 14 below shows that the green color signifies when to buy, and the red color signifies when to sell based on the positive or negative sentiments, as shown in fig. 13, of news headlines and titles from both the data frames. Positive news of a particular stock describes that the company will be doing good in the future and the best time to buy and hold the specific equity and vice versa for negative news.

\begin{figure}[h]
\centering
  \includegraphics[width=.22\textwidth]{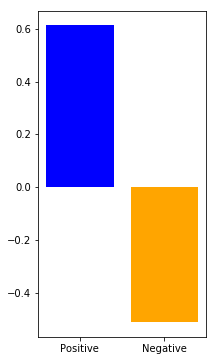}
\caption{Facebook Company's sentiments}
\label{fig:exampleFig}
\end{figure}

\begin{figure}[h]
\centering
  \includegraphics[width=.22\textwidth]{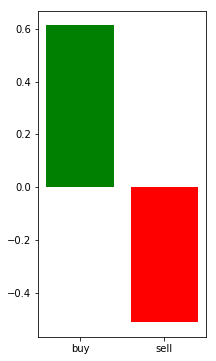}
\caption{Facebook stock's buy and sell decisions}
\label{fig:exampleFig}
\end{figure}

\noindent
Another focus of the study is the banking sector from the “Indian Titles” dataset. During this analysis, the companies included Axis Bank, HDFC, ICICI, Kotak Mahindra, SBI Bank, and IndusInd Bank. The compound score and positive or negative sentiments are calculated for each of the mentioned banking stocks depending on the news title, assisting in buying and selling. During this research, analysis was done on IndusInd bank.
\noindent
The analysis signifies when to buy or sell the IndusInd bank stock depending on the compound score as shown in fig. 15 The best compound score was close to 0.788, indicating that the stock for IndusInd Bank will go up in the future so, investors or traders should be thinking of buying the stock currently or holding the equity for some time. The worst compound score  -0.77, which is not a good time to buy since it is going down, and how far it will go down is difficult to forecast. Selling is another option if traders or investors are already holding the stock for a relatively long time. Fig. 16 shows buying and selling decisions to be made for IndusInd Bank.

\begin{figure}[h]
\centering
  \includegraphics[width=.22\textwidth]{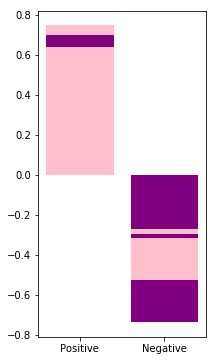}
\caption{IndusInd Bank's sentiments}
\label{fig:exampleFig}
\end{figure}

\begin{figure}[h]
\centering
  \includegraphics[width=.22\textwidth]{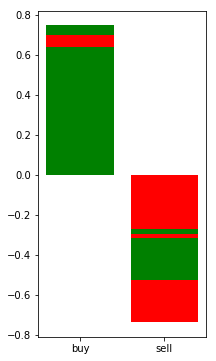}
\caption{IndusInd stock's buy and sell decisions}
\label{fig:exampleFig}
\end{figure}

\subsection{WordCloud for Global Headlines}
WordCloud is also known as the Tag Cloud. It is a cloud packed with loads of words or phrases in different dimensions, representing every word’s occurrence or significance. It is a graphical depiction of text data. Cloud words are generally singular words, and the importance of each word demonstrates different font sizes and  colors. WordCloud is extremely useful for discovering text data. A more significant word or phrase indicates greater weightage \cite{StandingScience}.
\begin{figure}[h]
\centering
  \includegraphics[width=.5\textwidth]{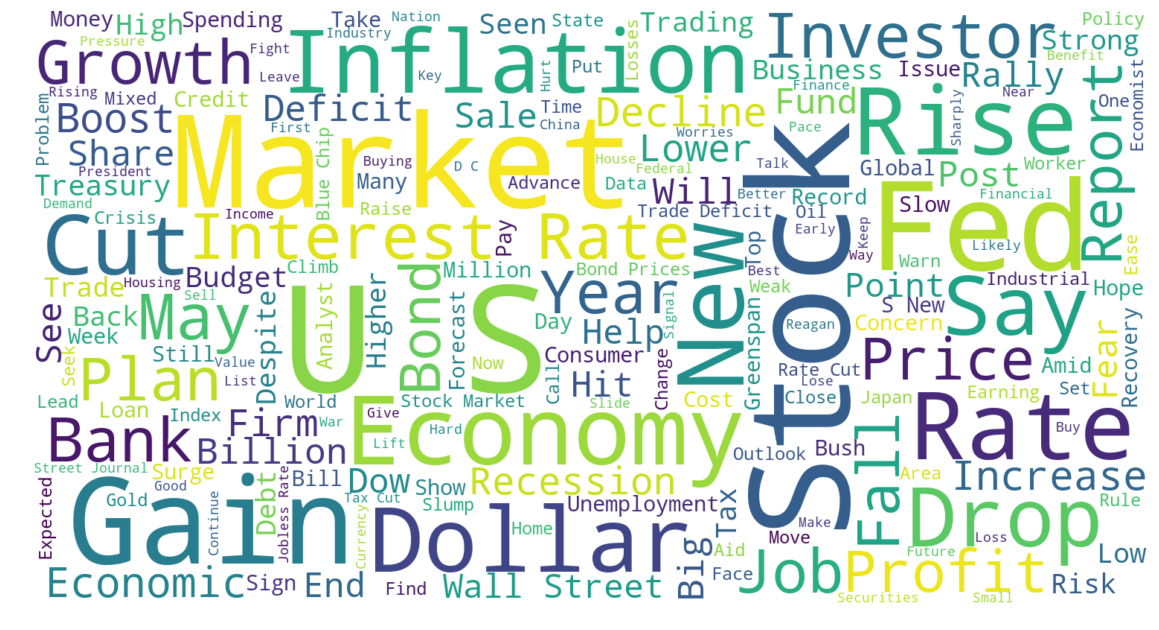}
\caption{WordCloud for Global headlines}
\label{fig:exampleFig}
\end{figure}
Fig. 15 displays the most frequent word explored in the “headline” column of the “Global Headlines” data frame. The top 10 most frequent words are as follows:
\begin{enumerate}
   \item Gain
   \item US
   \item Market
   \item Fed
   \item Stock
   \item Inflation
   \item Dollar
   \item Rise
   \item Economy
   \item Drop
\end{enumerate}

\subsection{WordCloud for Indian Titles}
DataFrame “Indian Titles” extracted from the Indian Financial News dataset has “Title” for which WordCloud has been designed to depict each word’s frequency or significance. The idea behind this Tag Cloud is to show
\begin{figure}[h]
\centering
  \includegraphics[width=.5\textwidth]{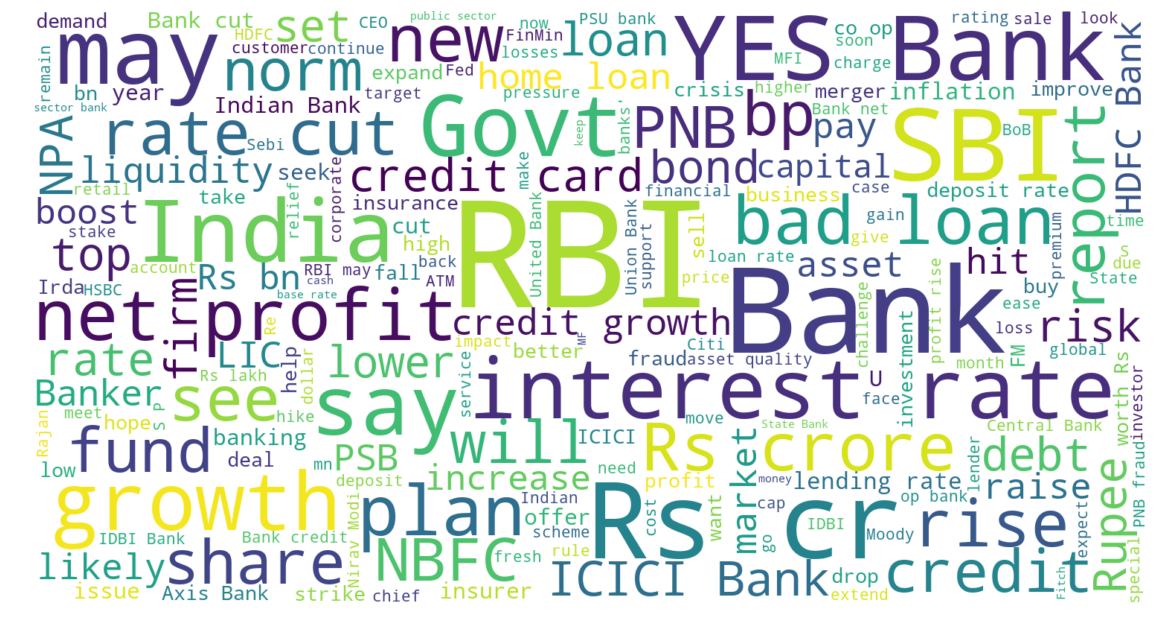}
\caption{WordCloud for Indian Titles}
\label{fig:exampleFig}
\end{figure}
how important these words are in the DataFrame. The more frequent these words or phrases appear, the more valuable they are since these are the top-notch tags in the Indian equity market.
Fig. 16 illustrates WordCloud of Indian Titles. The top 10 most frequent words or tags as follows:
\begin{enumerate}
   \item RBI
   \item Bank
   \item Rs cr
   \item SBI
   \item Yes Bank
   \item Interest Rate
   \item India
   \item Net Profit
   \item Govt.
   \item Fund
\end{enumerate}

\subsection{Distribution of sentiments for Global Headlines and Indian Titles}
This research analyzes the number of negative and positive sentiments for both the data frames. Opinions are established on the compound VADER score depending on the polarity of the news [89]. If a particular headline or title is optimistic, then the sentiment is positive (1), and if a specific story is gloomy, then sentiment class is negative (0).
\begin{figure}[h]
\centering
  \includegraphics[width=.4\textwidth]{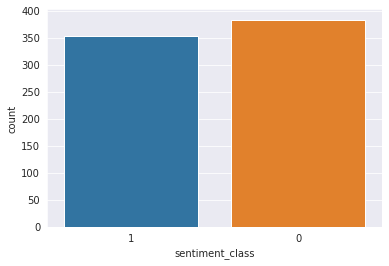}
\caption{Distribution of sentiments for Global Headlines}
\label{fig:exampleFig}
\end{figure}
As seen in Fig. 17, the number of negative sentiments is relatively higher than the number of positive emotions in Global financial news. The count of negative beliefs (0) is 375, as shown in Fig. 17, which is more than positive news (1), indicating a gloomy market that is not suitable for small traders or investors. However, significant investment banks or big investors can take positions by playing short in the market. Fig. 18 talks about the distribution of positive and negative sentiments for India Titles.
\begin{figure}[h]
\centering
  \includegraphics[width=.4\textwidth]{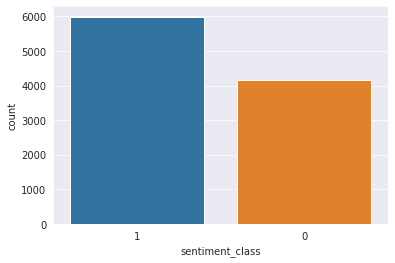}
\caption{Distribution of sentiments for Indian Titles}
\label{fig:exampleFig}
\end{figure}
\subsection{Distribution of sentence length for Global Headlines and Indian Titles}
In this research work, the focus is on the headlines and not the whole story. Through box plots \cite{PDFDatab}, outliers are checked both above the 90th percentile and below the 10th percentile. The reason is to get to know how long is the text, so it might get short during preprocessing that takes only imperative words in the sentence, helping in the model building. Fig. 19 and fig. 20 shows the distribution of sentence length for both the data frames. For Indian news titles, there are outliers in minimum and maximum sentence length. The max is more than 17.5, which is in strike contrast to less than 40 words in sentence length for Global headlines, and the minimum outlier is less than 2.5 for Indian news titles. 
\begin{figure}[h]
\centering
  \includegraphics[width=.4\textwidth]{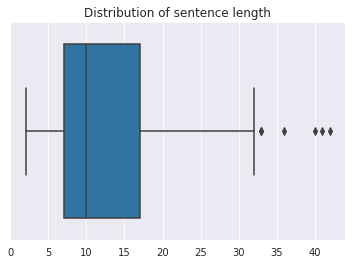}
\caption{Distribution of sentence length for Global Headlines}
\label{fig:exampleFig}
\end{figure}
\begin{figure}[h]
\centering
  \includegraphics[width=.4\textwidth]{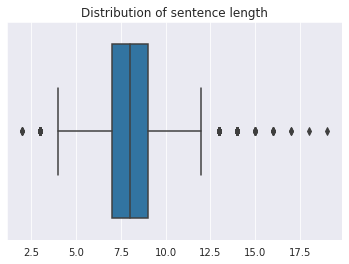}
\caption{Distribution of sentence length for Indian Titles}
\label{fig:exampleFig}
\end{figure}

\subsection{Distribution of positive headlines or negative headlines for Global Headlines and Indian Titles}
Financial news plays a vital role in the sentiments of investors. The moment the negative story of a particular company floats, the market reacts negatively. For example, FTSE 100 companies recover amidst Covid vaccine rollout, hoping that global growth will move forward. According to Bloomberg and BBC, on 4th January 2020, the US’s Nasdaq and Japan’s Nikkei 225 ended the year better than they began. However, FTSE 100 has yet to rebound since last January \cite{FTSENews}. The story of India is a bit surprising. Two days before the nationwide lockdown, the BSE Sensex had closed at 25,981 points, but now since the pandemic is still seething, the Sensex hit 42597.43 \cite{WhySlowdown}. Still, the markets are very much related to each other. Announcement of Oxford Vaccine in the United Kingdom and the Indian Government's approval to AstraZeneca  and Oxford University has led the news peaked for NSE Nifty 50 index and BSE Sensex \cite{IndianHindu}
\begin{figure}[h]
\centering
  \includegraphics[width=.5\textwidth]{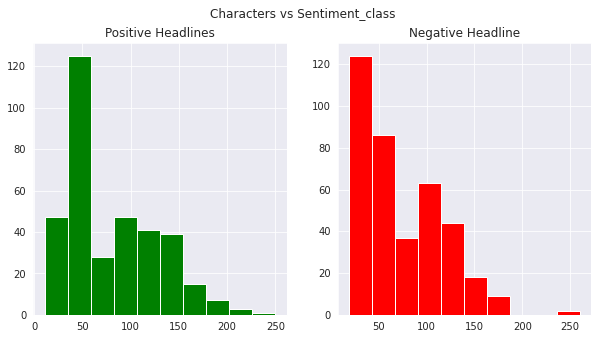}
\caption{Distribution of Positive and Negative headlines for Global headlines}
\label{fig:exampleFig}
\end{figure}
Fig. 21 and fig. 22 are the distribution of bullish and bearish news for both the data frames. The difference between the two distributions is that the Global headlines is positively skewed \cite{PositivelyFinance} and the "Indian Titles" is normally distributed \cite{NormalProperties}.
\begin{figure}[h]
\centering
  \includegraphics[width=.5\textwidth]{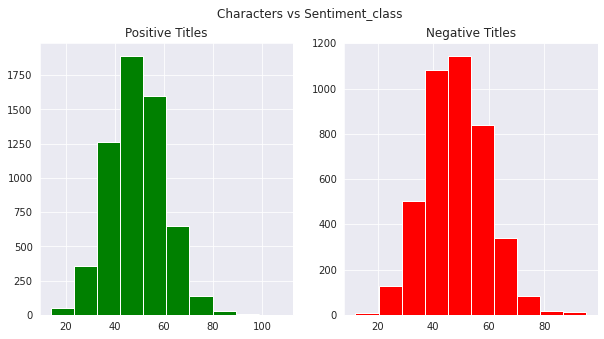}
\caption{Normally distributed Indian Titles}
\label{fig:exampleFig}
\end{figure}
\subsection{Distribution of stopwords for Global headlines and Indian Titles}
Stop words \cite{DroppingWords} are words that do not have that much importance in the sentence. Stopwords - the, yourself, that, etc. - can be disregarded while writing code in python without compromising the meaning of the sentence since they are already acquired in the corpus. So, these words were filtered out to help index and parse the text faster.

\section{Results and Discussions}
This section explores the findings of word embeddings, model building, and assessment metrics. The first segment of this section articulates word embedding techniques applied to various news headlines and titles. This section’s second segment focuses on model-building results such as RNN, and LSTM, an extension of RNN, implemented on the embedded news headlines or titles and discussed evaluating different modeling techniques and analyzing them for conclusions. In other words, the traits of the implemented models set the base for the inferences.
\begin{figure}[h]
\centering
  \includegraphics[width=.5\textwidth]{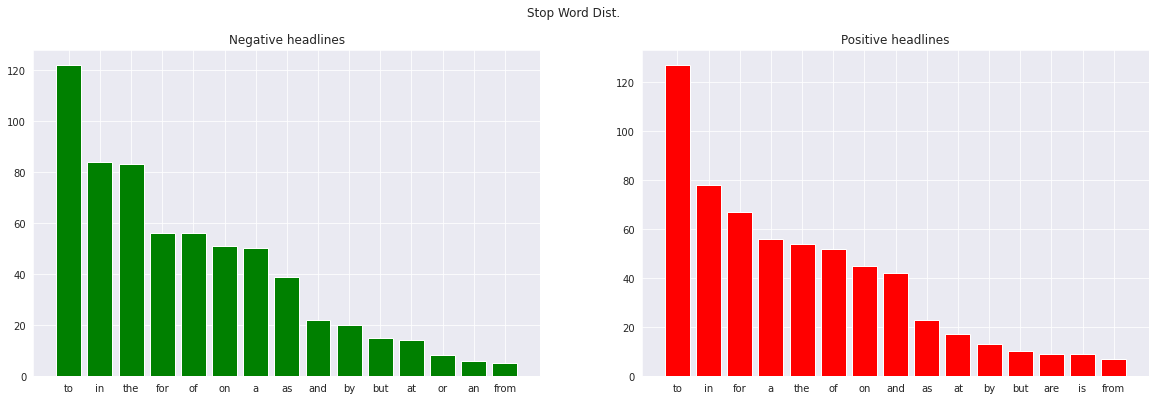}
\caption{Stopwords distribution for both Global and Indian titles}
\label{fig:exampleFig}
\end{figure}
\subsection{VADER-Valence Aware Dictionary and Sentiment Reasoner}
The algorithm implements VADER by calling the natural language toolkit (import NLTK) and VADER lexicon to calculate a sentiment score on news headlines or titles. These sentiment scores are known as compound scores that can be positive if the news is bullish and negative if the story is bearish. It is calculated by totaling every phrase's scores in the dictionary and then normalizing between -1 and +1.
\begin{figure}[h]
\centering
  \includegraphics[width=.5\textwidth]{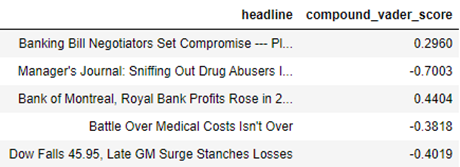}
\caption{Compound or Sentiment Score by VADER}
\label{fig:exampleFig}
\end{figure}
\subsection{Word Embedding}
The subsequent section explains the word embedding techniques in which phrases from the lexicon are charted to real number vectors. They characterize words or phrases that are critical developments of deep learning on perplexing NLP problems. The word embeddings do not comprehend the text as human beings would instead map the corpus's language's numerical structure.

\subsubsection{Bag of Words Model on Global headlines and Indian Titles}
Each time applying any algorithm in Natural Language Processing, it works on numbers. The text cannot be fed directly into the algorithm. Thus, the Bag of Words model preprocesses the text by changing it into fixed-length vectors that preserve the count of the most regularly used phrases. This method is called vectorization. For vectorization to work successfully, preprocessing of the data, such as converting the text to lower case, removing all non-word characters, and removing all punctuations, was done. Then vocabulary was determined and counted how many times each word or phrase appeared. Once fit, the tokenizer provided four attributes to query about the document learning. 
\begin{enumerate}
   \item Word counts: A dictionary of words and their counts.
   \item Word docs: An integer count of a total number of documents used to fit the tokenizer
   \item Word index: A dictionary of words and their uniquely assigned integers
   \item Document count: A dictionary of words and how many documents each appeared in the dictionary.
\end{enumerate}
There were a total of 8496 unique tokens for Global Headlines and 17989 tokens for India Titles. Once the tokenizer fitted the training data, it encoded the documents in the train and test datasets.

\subsubsection{TF-IDF Model on Global headlines and Indian Titles}
The previous BoW model counts the number of words in a document, but this model has some shortcomings:
\begin{enumerate}
   \item The term order is not reflected 
   \item The uniqueness of a term is not reflected
\end{enumerate}
To overwhelm these shortcomings, TF-IDF was applied.
TF-IDF stands for term frequency-inverse document frequency. TF-IDF figures out the significant words in a document. The first part of TF-IDF is term frequency (TF), which measures the number of times words appear in the document. It is the same as the Bag of Words. The second part is inverse document frequency (IDF), which measures the particular word's significance. The more documents the word appear in, the less significant that word is. In this process, the first CountVectorizer was imported from Sklearn and then instantiated as an object. CountVectorizer separated each sentence into tokens, and then fit-transform counted the number of times it occurred in a sentence. It created a list of vocabulary used for the vectorizer. It is the parameter that the vectorizer creates the matrix by using only input data or some other source.

\subsubsection{Word2Vec Model on Global Headlines and Indian Titles}
Before moving onto the Word2Vec model, the BoW and TF-IDF models had some limitations:
\begin{enumerate}
   \item Unable to capture the meaning of the text: The basic implementation of BoW and TF-IDF doesn’t account for the order in which words appear in a particular document. Often stop words are deleted from the dictionary, which may carry important information with them. The algorithms capture meaning to some extent by forming bigrams and further n-grams from the dictionary, but this again results in resource hogging.
   \item Dimension: While it is inevitable that not every document has an equal number of unique words available. Using BoW, how are the documents compared? The answer  lies in TF-IDF. In this, every doc has a vector representation of the same dimension. To do so, calculate the total number of unique words in all documents combined and repeat the similar vectorizing procedure for each document. In this way, vectorized representation of documents has many zero entries due to those words that are not present in the particular document in consideration. At the same time, this solution works if dealing with a few documents. With more documents and especially if they are from diverse backgrounds, lots of computational resources are wasted. The curse of dimensionality starts hitting hard, a common problem in machine learning.
\end{enumerate}
The Word2Vec model can overcome the limitations of BoW and TF-IDF. Word2Vec utilizes two methods (CBOW and Skip-Gram) to produce a vectorized form of words, both using a two-layer neural network architecture at their core, making the Word2Vec model a much more sophisticated model than BoW and TF- IDF. This process helps in limiting the dimension of vectorized representation of words/documents. 
These vectors do carry meaning with them because closely related words are positioned closely in the vector space. For example, the word pair ‘terrible’ and ‘horrible’ would result in a similar word embedding. While the word pair ‘creative’ and ‘unimaginative’ would result in different vector representations. The thought process behind this is that more similar words should have a high similarity score instead of other words. 

\noindent
The similarity between any two vectors is calculated by taking the two vectors' dot product, given that they are normalized. It is nothing but the famous cosine similarity. This research work employed the Word2Vec model by using Google’s pre-trained word2vec model. In the Indian Titles, the column “Title” is transformed into a list of lists to start training with the Word2Vec model. 

\noindent
The function “simple-preprocess” of the GenSim package is used to transform each title’s row into a set of tokens. For example, “Apple is the best stock.” Its tokenized version - “Apple,” “is,” “the,” “best,” “stock.” The same is applied to column “headline" for "Global headlines". Once tokens are created, the Word2Vec model constructs the vocabulary by retrieving all unique words from sub-lists of “documents,” then the Word2Vec model is specified. The studied vocab of tokens is stored in ‘.wv’ for creating vectors.

\noindent
There are some features of the Word2Vec model:
\begin{enumerate}
   \item Similar words
   \item Given words A and B, find other words that are similar to A and opposite to B.
\end{enumerate}
\subsubsection{FastText Model on Global Headlines and Indian Titles}
Word2Vec considers each phrase in the collection and creates a vector for each one. On the other hand, FastText, a branch of the word2vec model, handles each word as an organized character of n-grams. For instance, the word vector “orange” is a total of the vectors of the n-grams “<or,” “ora,” “oran,” “orange>,” “ran,” “rang,” “ange>,” “nge,” “nge>” (presuming hyperparameters for smallest ngram[minn] is 3, and biggest ngram[maxn] is 6).
\begin{figure}[h]
\centering
  \includegraphics[width=.5\textwidth]{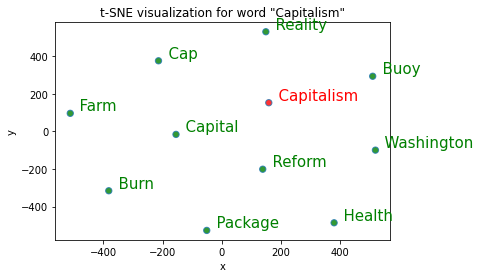}
\caption{t-SNE plot for word “capitalism” for Global Headlines.}
\label{fig:exampleFig}
\end{figure}
The t-SNE plot shows the most similar words to capitalism. As shown in Fig 25, Buoy, Capital, and Reform are very close to Capitalism, indicating similarity. The same goes for Indian Titles. Here, we have taken words such as “investment” and “investor” to get the most similar words, similarity scores, and TSNE plot \cite{AnScience}.
\\
\noindent
The t-SNE plot shows the most similar words to “Investment.” As shown in Fig. 26, “Divestment” and “Placement” are very close to Investment, indicating similarity.
\begin{figure}[h]
\centering
  \includegraphics[width=.5\textwidth]{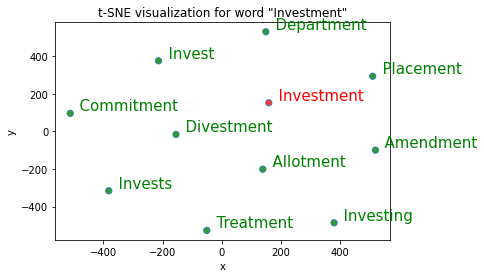}
\caption{t-SNE plot for word “investment” for Indian Titles.}
\label{fig:exampleFig}
\end{figure}
\subsubsection{GloVe Model on Global Headlines and Indian Titles}
GloVe, an unsupervised learning algorithm, acquires vector characterization or depiction for words. GloVe permits to collect text and instinctively transmutes each word in that compilation into a stance of lofty dimensional space, which means related terms positioned together. Before glove, the methods of word characterizations were divided into two segments, LDA (Latent Dirichlet allocation) [98] and Word2Vec. LDA generates less than average dimensional vector by SVD, whereas Word2Vec applies three levels neural network to perform word pair classification task. The significant part from Word2Vec is that related words are placed together in the vector space, and vectors' mathematical operations can cause semantic connections; However, LDA cannot sustain such relationships. The incentive behind GloVe is to push the model to understand linear relationships developed on a conjunction matrix. Most importantly, GloVe is a hybrid algorithm that uses ML established on a statistic matrix, a distinct difference between GloVe and Word2Vec.
\subsubsection{BERT Model on Global Headlines and Indian Titles}
BERT’s results show that a bi-directionally trained language model can have a more profound sense of flow and context language than single-direction language models. The BERT model includes a novel technique named Masked Language Model (MLM), which allows bidirectional training in models. The BERT makes use of a transformer, an attention mechanism that learns contextual relations between words or sub-words in a text. BERT is an NLP tool developed by Google. It makes use of the textual data which is available on the web. Word2Vec and older models have been applied in a context-free manner, but the BERT model is trained in a bidirectional way, making BERT aware of the context of the word \cite{OmidvarLearningHeadlines}. For example, BERT can differentiate between a bucket list and a bucket full of water. In this research work, BERT converts the news headlines or titles to fixed-length vectors. The steps followed are:
\begin{enumerate}
   \item Import the BERT client module and initialize a variable
   \item Call encode method of BERT client to get the fixed-length vector
\end{enumerate}

\begin{figure}[h]
\centering
  \includegraphics[width=.5\textwidth]{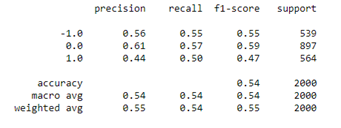}
\caption{Classification score of BERT model}
\label{fig:exampleFig}
\end{figure}
\noindent
The classification report shows the model's accuracy, which is just 54\texttt{\%} and not that good. The precision, recall, and f1-score are also not up to the mark. This model doesn’t perform its best as planned.

\section{Model Building}
The focus of this research is laid on building deep learning models on the above embedding techniques applied to the data frames extracted from both the Indian Financial News and Full Economic News.

\subsection{RNN Vs LSTM for Indian Titles and Global Headlines}
RNN is intended to operate with sequential data, and this data is mostly in the form of text, video, audio, etc. However, in this work, the focus is on the news title or headlines (text). RNN utilizes the earlier data in succession to produce the recent outputs \cite{Yu2019AArchitectures}. One of the drawbacks of RNN is the Vanishing Gradient problem,  a reason for the short-term memory problem as RNN practice more steps \cite{SherstinskyFundamentalsNetwork}.
\\\\
\noindent
A specialized version of RNN is created to overcome this problem, known as LSTM (Long Short-Term Memory). LSTM uses memory cells to keep the activation value of earlier words in lengthy sequences. As RNN hasn’t got good results because of the vanishing gradient issue on epochs of 10, 50, and 100, the results were not appropriate for whatever epochs were set. So, the research implemented LSTM, a version of RNN resolving vanishing gradient \cite{VanHoudt2020AModel}. The Figure shows an illustration of Simple RNN with epoch 50.
\begin{figure}[h]
\centering
  \includegraphics[width=.5\textwidth]{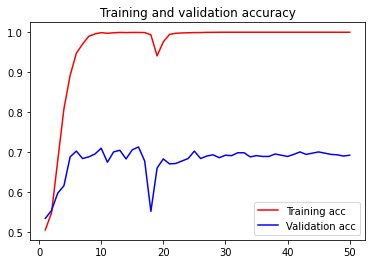}
\caption{Simple RNN Accuracy on epoch 50}
\label{fig:exampleFig}
\end{figure}
\begin{figure}[h]
\centering
  \includegraphics[width=.5\textwidth]{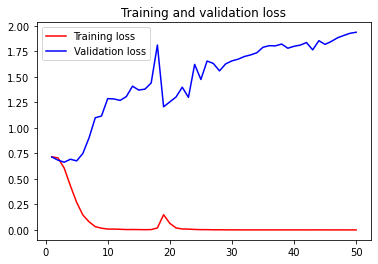}
\caption{Simple RNN Loss on epoch 50}
\label{fig:exampleFig}
\end{figure}
\noindent
The accuracy for Simple RNN is 69.56\texttt{\%} that is in strike contrast with LSTM models, which has better accuracy relative to Simple RNN.

\subsection{Building GloVe word embedding with LSTM model on Indian Titles}
The pre-trained Word Embedding GloVe provides more understandings for phrases, and it is very challenging to build a corpus of huge words, so the assistance of Standford’s GloVe embedding is used. For this study, Standford’s GloVe Embedding generates an index of phrases or words charted to identified embeddings by analyzing the data dump of pre-trained embeddings, then load word or phrase embeddings into an embeddings\texttt{\_}index. A total of 4,00,000 GloVe pre-trained word vectors were created \cite{Pennington2014GloVe:Representation}. While developing an LSTM model, one should keep in mind essential things such as model architecture, hyperparameter tuning, and model performance. In the word cloud, some phrases predominantly feature positive and negative news titles. It could be a problem if researchers implement a machine learning model such as Naïve Bayes, SVD, etc., so that’s why the study implements sequence models with two layers \cite{Sari2020TextFeatures}.
\\
\noindent \\
The following are the steps taken during the process:
\begin{enumerate}
   \item Build the embedding layer
   \item Specifies the maximum input length to the Embedding layer
   \item Utilize the output from the last embedding layer that yields a 3-D tensor into the LSTM layer.
   \item Setting the dropout layer to 50\texttt{\%} instead of lower since an increase can give better results.
   \item Setting the dense layer to generate an output dimension of 512 is essential.
   \item Finally, the output is fed into a “sigmoid” layer.
   \item After the model training, evaluate the model’s performance
\end{enumerate}
\begin{figure}[h]
\centering
  \includegraphics[width=.5\textwidth]{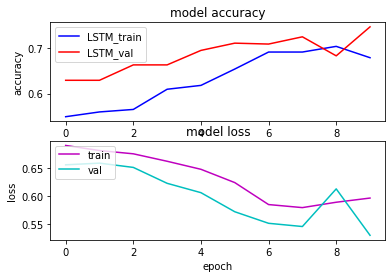}
\caption{Learning Curve of loss and accuracy for Indian Titles}
\label{fig:exampleFig}
\end{figure}
The range of prediction scores is between 0 and 1. Two classes are classified by defining a threshold value for them. In this case, the threshold value of 0.5 is set during coding. If the score is above it, then it is classified as a positive sentiment.
\\
\noindent \\
Around 75\texttt{\%} accuracy is good enough considering the baseline human accuracy. This model is good enough to handle most tasks for sentiment analysis. Precision for negative is 73\texttt{\%} and for positive is 76\texttt{\%}, recall for negative is 51\texttt{\%} and for positive is 89\texttt{\%}, and F1-score for negative and positive is 60\texttt{\%} and 82\texttt{\%}.

\begin{figure}[h]
\centering
  \includegraphics[width=.5\textwidth]{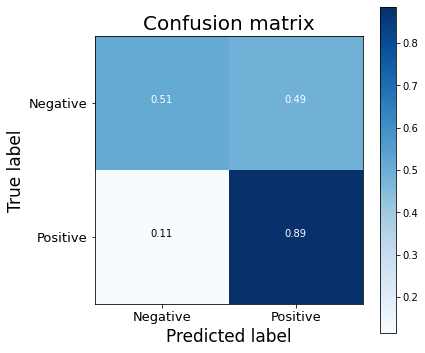}
\caption{Confusion matrix of GloVe on Indian Titles}
\label{fig:exampleFig}
\end{figure}

\begin{figure}[h]
\centering
  \includegraphics[width=.5\textwidth]{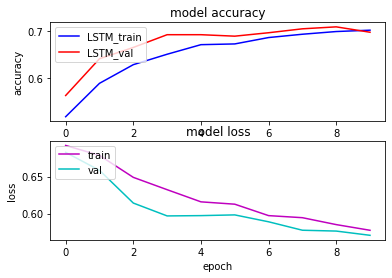}
\caption{Learning Curve of loss and accuracy for Global Headlines}
\label{fig:exampleFig}
\end{figure}

\subsection{Building GloVe word embedding with LSTM model on Global Headlines}
The process for Global\texttt{\_}headlines\texttt{\_}df is similar to section 13.2 on Indian\texttt{\_}Title\texttt{\_}df. GloVe embedding is the same, and no changes while implementing the LSTM model; However, the outcomes are quite different. Fig. 32 shows both LSTM\texttt{\_}train and LSTM\texttt{\_}val cross each other somewhere close to 70\texttt{\%}. The loss curve for train and val is parallel, but both are going down. This model’s accuracy for global\texttt{\_}headlines\texttt{\_}df is 70\texttt{\%}, which is relatively less than the same model for Indian\texttt{\_}titles\texttt{\_}df, 75\texttt{\%} as shown in Fig. 31. However, one cannot say that this model is not better. The two datasets are different. One is focusing on Indian news and the other on global reporting.

\subsection{LSTM model on BoW for Global Headlines}
Here, the number of unique tokens is 6612. Maximum words used 10,000. Pad sequences truncate the words that cross the threshold of 50.

\begin{figure}[h]
\centering
  \includegraphics[width=.4\textwidth]{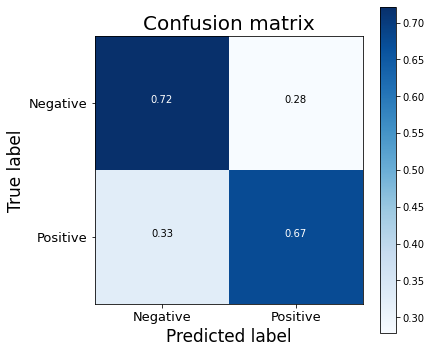}
\caption{Confusion matrix of GloVe on Global headlines}
\label{fig:exampleFig}
\end{figure}
\begin{figure}[h]
\centering
  \includegraphics[width=.4\textwidth]{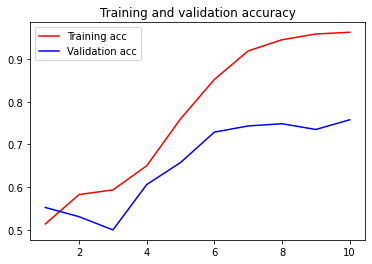}
\caption{Training and Validation Accuracy for Global headlines}
\label{fig:exampleFig}
\end{figure}
\noindent
The Fig. 34 and fig. 35 shows how the model performs. There is a considerable difference or a gap between training accuracy and validation accuracy, around 25\texttt{\%}, but the model accuracy is above 75\texttt{\%}. At the beginning of the model loss, there is a little difference between training and validation, but later we see a massive difference.
\begin{figure}[h]
\centering
  \includegraphics[width=.4\textwidth]{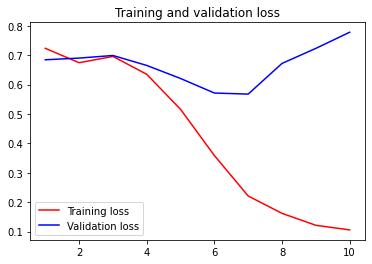}
\caption{Training and Validation Loss for Global headlines}
\label{fig:exampleFig}
\end{figure}

\subsection{LSTM model on BoW for Indian Titles} 
The validation accuracy, approx 90\texttt{\%}, on Indian\texttt{\_}Title\texttt{\_}df is better than Global\texttt{\_}headlines\texttt{\_}df, but since the data frames are analyzed separately, it doesn’t matter. However, it is better than simple RNN.
\begin{figure}[h]
\centering
  \includegraphics[width=.4\textwidth]{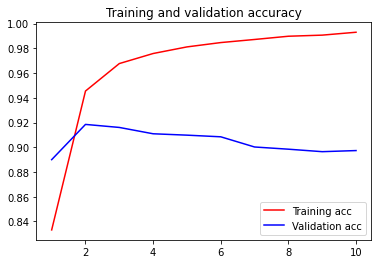}
\caption{Training and Validation accuracy for Indian Titles}
\label{fig:exampleFig}
\end{figure}
\begin{figure}[h]
\centering
  \includegraphics[width=.4\textwidth]{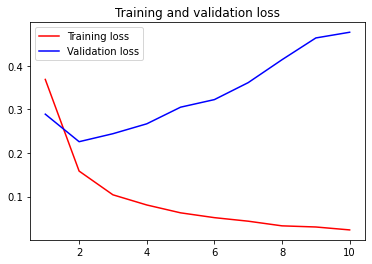}
\caption{Training and Validation loss for Indian Titles}
\label{fig:exampleFig}
\end{figure}

\subsection{LSTM model on TF-IDF for Indian Titles and Global Headlines}
\begin{figure}[h]
\centering
  \includegraphics[width=.38\textwidth]{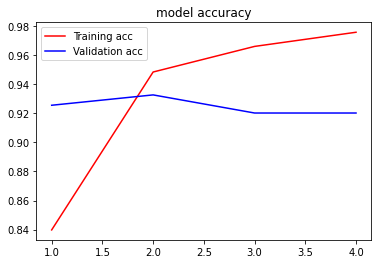}
\caption{Model accuracy for Indian Titles}
\label{fig:exampleFig}
\end{figure}

\begin{figure}[h]
\centering
  \includegraphics[width=.4\textwidth]{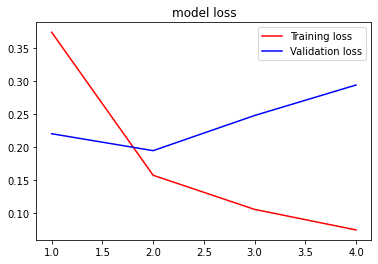}
\caption{Model loss for Indian Titles}
\label{fig:exampleFig}
\end{figure}
\noindent
The validation accuracy for the LSTM model on Indian\texttt{\_}Title\texttt{\_}f is 93.27\texttt{\%} and model loss is 29.38\texttt{\%}.
\\ 
\noindent \\
In the classification score of Indian Financial News, the accuracy is 91\texttt{\%}. It also gives precision -  91\texttt{\%}, recall 94\texttt{\%}, and f1-score 93\texttt{\%} for positive sentiments and precision -  91\texttt{\%}, recall 87\texttt{\%}, and f1-score 89\texttt{\%} for negative sentiments.
\\
\noindent \\
The validation accuracy for Global\texttt{\_}headlines\texttt{\_}df is 81.86\texttt{\%}, and validation loss is approx. 60\texttt{\%}.
In the classification score, the precision is 82\texttt{\%} for negative and 85\texttt{\%} for positive, recall is 83\texttt{\%} for negative and 84\texttt{\%} for positive, f1-score is 83\texttt{\%} for negative and 84\texttt{\%} for positive. The overall accuracy is 84\texttt{\%}.

\begin{figure}[h]
\centering
  \includegraphics[width=.4\textwidth]{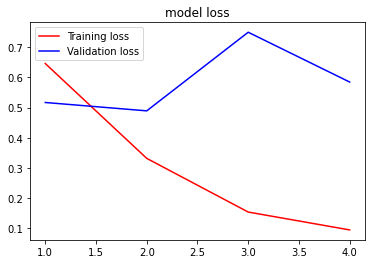}
\caption{Model loss for Global headlines}
\label{fig:exampleFig}
\end{figure}

\subsection{LSTM model on Word2Vec for Indian Titles and Global Headlines}
Here,  the dataset for binary classification included 11,220 news titles for training and 11,220 for testing. The news title was first encoded so that each word or phrase is represented by a unique integer. Word padding made all the vectors of the same length (max\texttt{\_}length). The embedding layer required the specification of the vocabulary size (vocab size) and the size of the real-valued vector space EMBEDDING\texttt{\_}DIM = 100. Now, ready to define the neural network model. Word embedding is a beautiful approach, but there is another approach to build a sentiment classification model. Instead of training the embedding layer, we first separately learned word embeddings and then passed them to the embedding layer. We used the Gensim implementation of Word2Vec. The first step was preprocessing; then, the word2vec algorithm processed documents sentence by sentence. Once the model was trained on the news title dataset, it built a vocabulary size of 11027. The next step was to convert the word embeddings into a tokenized vector. The integer encoded titles were mapped to the index of a specific vector in the embedding layer. Therefore, it was essential to lay the vectors out in the embedding layer, encoded words mapped to the correct vector. Now embeddings were mapped from the loaded word2vec model for each word to the tokenizer\texttt{\_}obj.word\texttt{\_}index vocabulary and created a matrix with word vectors.
\begin{figure}[h]
\centering
  \includegraphics[width=.4\textwidth]{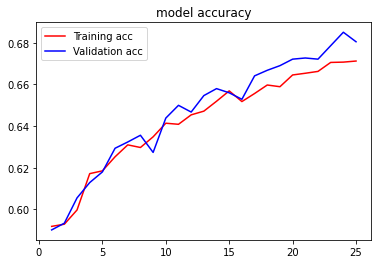}
\caption{Model accuracy for Indian Titles}
\label{fig:exampleFig}
\end{figure}

\begin{figure}[h]
\centering
  \includegraphics[width=.4\textwidth]{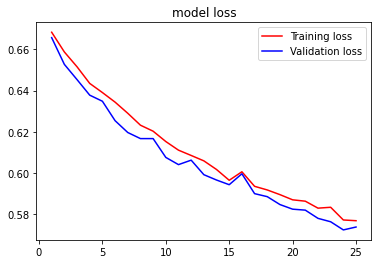}
\caption{Model loss for Indian Titles}
\label{fig:exampleFig}
\end{figure}
\noindent
Then, the trained embedding vectors were ready to be used directly in the embedding layer. To train the sentiment classification model, VALIDATION\texttt{\_}SPLIT=0.2. There was no specific reason for setting at two but setting this number gave good results.
\\
\noindent\\
The Fig. 42 and fig. 43 shows that the model after each epoch is improving the model accuracy. After a few epochs, validation accuracy is close to 69\texttt{\%}. The plot is perfect, parallel to training accuracy and training loss.
\begin{figure}[h]
\centering
  \includegraphics[width=.4\textwidth]{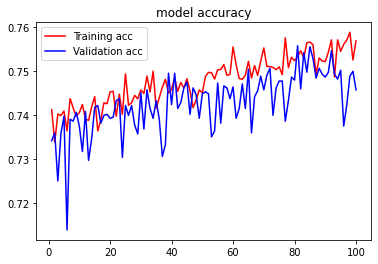}
\caption{Model Accuracy for Global Headlines}
\label{fig:exampleFig}
\end{figure}
\begin{figure}[h]
\centering
  \includegraphics[width=.4\textwidth]{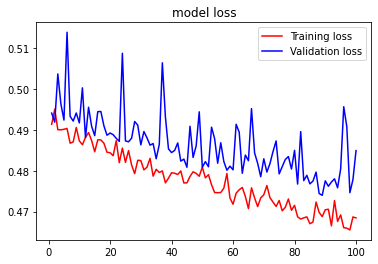}
\caption{Model loss for Global Headlines}
\label{fig:exampleFig}
\end{figure}
\\
\noindent \\
The fig. 44 and fig. 45 shows the validation accuracy for Global\texttt{\_}headlines\texttt{\_}df, which is 74\texttt{\%} that is less than training accuracy but parallel to training accuracy on epoch = 100. Validation is close to 49\texttt{\%}, similar to training loss but way above it.

\subsection{LSTM model on FastText for Indian Titles and Global Headlines}
\begin{figure}[h]
\centering
  \includegraphics[width=.4\textwidth]{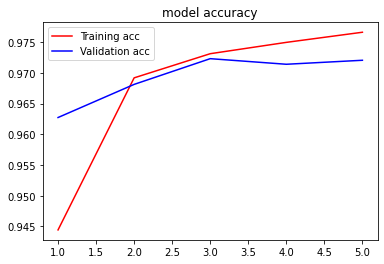}
\caption{Model accuracy of FastText on Global Headlines and Indian Titles}
\label{fig:exampleFig}
\end{figure}
\noindent
Like in the above sections, TensorFlow/Keras is used to implement Long Short Term Memory (LSTM) neural network and utilizes self-trained embeddings, i.e., FastText. As the most other sections, this section was much more computationally demanding, so running it on a CPU was very frustrating as it felt like ages. So, it was implemented on Google Colab as it offers free TPU usage. As before, the data was split into a training and testing set for cross-validation. In dealing with feature creation, tokenizer function in Keras was used. Each news title or headline was transformed into a vector of integers in which each element represented a token. Defining callback function as it does two things – it stops the training if there is no progress, and it saves model checkpoints after each epoch. Finally, the model was trained. After some iterations, the activity reached its best performance.
\begin{figure}[h]
\centering
  \includegraphics[width=.4\textwidth]{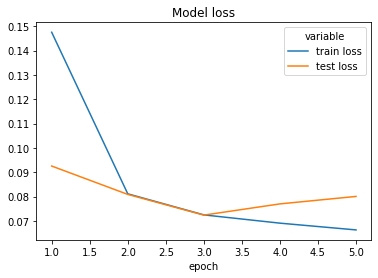}
\caption{Model loss of FastText on Global Headlines and Indian Titles}
\label{fig:exampleFig}
\end{figure}
The validation accuracy is close to 97\texttt{\%}. While training loss keeps decreasing through all epochs, the test score stops falling and slightly increases after six epochs. Thus, it was wise to stop the training here as going on would only lead to overfitting our model. As shown in Fig. 56, F1-score for macro avg is 92\texttt{\%}. A lot of parts and pieces have already been tweaked in this model. So, the best accuracy was achieved at 97\texttt{\%}.

\section{Conclusions}
This section discusses the conclusions and recommendations derived from the research employed on the news headlines dataset, and how deep learning models RNN and LSTM perform on both the Indian Financial News [106] and Global Financial Headlines. Outcomes are also discussed in this section to know the best-performing model. During model building, it was noticed that the enhanced deep learning LSTM model performed better. The table below displays the comparison of models with their accuracy.
\begin{table}[!h]
\centering
\begin{tabular}{| c | c |} 
\hline
\textbf{Models} & \textbf{Accuracy} \\ [0.5ex]
\hline\hline
RNN & 69.56\texttt{\%} \\
\hline
GloVe with LSTM1 & 75\texttt{\%} \\
\hline
GloVe with LSTM1 & 70\texttt{\%} \\ 
\hline
BoW with LSTM1 & 75\texttt{\%} \\ 
 \hline
BoW with LSTM2 & 90\texttt{\%} \\ 
\hline
TF-IDF with LSTM1 & 91\texttt{\%} \\ 
\hline
TF-IDF with LSTM2 & 84\texttt{\%} \\
\hline
Word2Vec with LSTM1 & 72\texttt{\%} \\  
\hline
Word2Vec with LSTM2 & 74\texttt{\%} \\  
\hline
FastText with LSTM & 97\texttt{\%} \\  
\hline
BERT Model & 54\texttt{\%} \\  [1ex]
\hline
\end{tabular}
\caption{Comparison of the models}
\label{table:ta}
\end{table}

The key conclusions in this research:
\begin{enumerate}
   \item Deep learning LSTM outperforms the RNN model. Its performance is steady across all word embedding models.
   \item Because of the vanishing gradient problem, the RNN model was not the best performer.
   \item Out of the phrase embedding techniques with LSTM, FastText with LSTM model is the best performing model with an accuracy of 97\texttt{\%}.
   \item During the model building phase, BoW with LSTM was supposed to be the worst performing model, but it came out to be one of the best performing models; however, it remained short of TF-IDF and FastText with LSTM.
   \item Word2Vec with LSTM disappointed, although had high hopes.
   \item BERT is itself a model. It has the lowest accuracy of 54
 \end{enumerate}

\section{Answering Research Questions}
\begin{enumerate}
   \item Comparison between different sectors (such as Oil and Gas, Banking, IT, etc.) of the economy concerning the financial news and which one performs better?
   \\
   \noindent
   The COVID-19 news has been dramatically disruptive in the financial activity of almost all countries. Most of the sectors have been badly impacted as the demand plunged, but higher progress was observed with some prominent exceptions. The oil and Gas, Travel and Tourism, and Hospitality industry have been slowed down because of the complete lockdown across the countries, affecting the demand for traveling, staying, and using oil and gas. Pharmaceuticals, Telecom, and Food and Agriculture have been growing. Mainly for India that has been exporting Hydroxychloroquine to Middle-East, UK, US, Canada, etc. The telecom sector has been continued to run since the “work-from-home” policy got implemented, noticing a surge in internet connections. Agriculture is the foundation of many countries, and its impact has been less since several Governments announced the free movement of essentials such as fruits, vegetables, milk, etc.
   \\
   \item Which company is the best and the worst performer in its sector concerning the financial news?
   \\
   \noindent
   As eating out was not an option during the Coronavirus pandemic, takeout and delivery such as DoorDash, GrubHub, Zomato, Swiggy, etc., won customers. Brick-and-mortar retailers lost to online retailers as customers who loved to go to malls could not do so. Amazon continued to be vast and powerful as it received a big boost because shoppers could not go shopping. Zoom, a web conferencing company, became a leader during covid-19. Other companies such as Cisco and GoToMeeting were also doing well. As people stopped eating out, it became clear to make their dinners. So, supermarkets remained open for essentials. Prices increased as people were scared of scarcities, and demand went up. Supermarkets also didn’t have to rely on coupons and price promotions to get the customers in, turning them into profit. Massive demand for personal hygiene products assists the growth. Reckitt Benkiser, Himalaya Global Holdings Ltd. Unilever Plc., Proctor and Gamble, etc., gained traction. While theatres or cinemas went off, Netflix added approx. Sixteen million new users during worst times.
   \\
   \item How does positive sentiment lead to buying, negative emotions lead to selling, and neutral leads to nothing?
   \\
   \noindent
   VADER, a lexicon sentiment analysis tool, is used to calculate the compound score on financial news. It provides a Valence Score to the words by observation rather than logic. The terms ‘bad’ and ‘hate’ are gauged negatively, and ‘good’ and ‘love’ are judged positively. VADER is measured in the range of -4 to +4.
   \\
   \item Are these indexes related to each other, i.e., does one affects the other or not?
   \\
   \noindent
   Financial markets around the globe have volatility mainly led by news or incidents across the world. For example, news on the recession, increase in commodity prices, variation in crude oil prices, etc., are some of the primary reasons why one index is affected by another index or established on the advancement of global markets. Markets are getting more mingled with each other. Analysts pursue and debate global events and movements very closely.
   \\
   \item Is there a way out when global news does not affect the local indices such as (nifty 100) or does the news worldwide impact all indices?
   \\
   \noindent
   Most of the time, news around the globe impacts local indices, but when the local consumption is high, then local indices don’t rally behind the global market, depending on the situation. The export market is driven by external demand. In such a case, the multinationals have to look inner markets for growth, helping local indices grow stronger.
   \\
   \item Which index, such as nifty 100, S and P 500, FTSE 100, etc., performs better with the Global Financial News, helping investors where to invest?
   \\
   \noindent
   One of the most complicated questions above is that a good policy change for one country might worsen for another. An increase in global crude oil prices is good for OPEC countries but might not be suitable for developing countries. FTSE 100 comprises businesses whose income is derived overseas. China, the biggest  buyer of natural resources, has caused a decrease in the price of commodities that these kinds of companies sell to them since the financial crises, contributing to less profitable earnings and slowing stock prices to go up. This event surrounding the British Banking Sector led to the poor performance of the FTSE 100.
   \\
   \item Which Deep learning model – RNN, or LSTM, performs better on sentiment analysis?
   \\
   \noindent
   After analyzing, FastText with LSTM has approx. 97\texttt{\%} accuracy that is better than the other models. Although RNN was intended to perform better, it has the lowest accuracy of 69.56\texttt{\%}.
   \\
   \item Comparing the word embedding methods – Word2Vec, GloVe, BERT, TF-IDF, Bag of Words, and FastText - which one is the best?
   \\
   \noindent
   Though BoW and TF-IDF tally the number of terms in a document, it has significant inadequacies - the word order and uniqueness of a term are not contemplated, incapable of catching the sense of the writing and dimensions. To overwhelm these inadequacies, Word2Vec(skip-gram) is employed. Word2Vec is good as it studies each word in the corpus and creates a vector for each word; however, it overlooks that some words occur more frequently than others. In contrast, FastText takes each phrase as an ordered character of n-grams, and GloVe stresses to carry additional information. It builds the word embeddings so that a mixture of word vectors connects directly to the chance of these words’ occurrences in the dictionary. BERT has its advantages. It predicts the neglected words in the context as it examines  every sentence or phrase with no particular course of direction. It plays a more superior role in grasping the sense of words than former NLP techniques.
 \end{enumerate}
 
\section{Future Recommendations}
The method and analysis adopted in this research kindly justified the objectives. However, some topics are there where more study can be applied. The subsequent lines explain the essential suggestions in both intellectual and business-related contexts.
\begin{enumerate}
    \item Simple RNN is a deep learning model applied to Indian financial news and Global Full Economic headlines datasets. However, the model's effectiveness has been mediocre than the Long Short Term Memory (LSTM) model's effectiveness. This shortcoming is  attributed to the Vanishing Gradient problem. More research is needed to understand the nuances of Simple RNN and discover the limitations and possibility of improvements.
    \\ 
    \item In this research, various word embedding techniques are applied, including the BERT model. It didn’t perform as hoped. A critical analysis is needed on how BERT can be improved. For future research, Skip-gram is another word embedding methodology that can give more insights if used with both machine learning and deep learning methods.
    \\
    \item During this study, LSTM and Simple RNN has been applied to the word embedding techniques. For future work, Generative Adversarial Networks (GANs) or Radial Basis Function Networks (RBFNs) should be combined with the above word embeddings to get a better state of the art working.
    \\
    \item As part of this research, top companies of various sectors from the Global indices such as S and P 500, FTSE 100, Nifty 100, etc. were studied with its news headlines to understand and analyze the companies' sentiments established on which buy and sell decisions of a particular stock seem promising. For further research work, the news sentiments of a specific company can be merged with its stock price data to predict the company’s open or closing price.
    \\
    \item There has been significant research in reinforcement learning for predicting the change in the stock price. Research in gauging and distinguishing the reinforcement algorithms by evaluating them with deep learning algorithms will increase the knowledge base and profitable applications. 
\end{enumerate}

\begin{itemize}
  \item \textbf{Funding} This research has not been funded by \noindent any agency. 
  \item \textbf{Conflicts of Interest} The authors declare that \noindent they have no conflict of interest
  \item \textbf{Availability of data and material} Not applicable
  \item \textbf{Code availability} Not applicable
\end{itemize}

\bibliographystyle{unsrt}
\bibliography{referencessenti}

\begin{thebibliography}{100}

\bibitem{APTwitter}
{AP CorpComm on Twitter: "Advisory: @AP Twitter account has been hacked. Tweet
  about an attack at the White House is false. We will advise more as soon as
  possible." / Twitter}.

\bibitem{FastTextRepresentation}
{FastText | FastText Text Classification {\&} Word Representation}.

\bibitem{Samuels2019SentimentReviews}
Antony Samuels and John Mcgonical.
\newblock {Sentiment Analysis on Customer Feedback Data: Amazon Product
  Reviews}.
\newblock 2019.

\bibitem{Rajput2020WordPandemic}
Nikhil~Kumar Rajput, Bhavya~Ahuja Grover, and Vipin~Kumar Rathi.
\newblock {Word frequency and sentiment analysis of twitter messages during
  Coronavirus pandemic}.
\newblock 4 2020.

\bibitem{UnderstandingScienceb}
{Understanding Political Twitter. Using tweet sentiment analysis to{\ldots} |
  by Duncan Grubbs | Towards Data Science}.

\bibitem{10Data}
{(10) (PDF) Analyzing Political Sentiment using Twitter Data}.

\bibitem{Statista}
{• Assets managed by hedge funds globally 1997-2019 | Statista}.

\bibitem{Sahu2020SupervisedTechnique}
Kalyan Sahu, Yu~Bai, and Yoonsuk Choi.
\newblock {Supervised Sentiment Analysis of Twitter Handle of President Trump
  with Data Visualization Technique}.
\newblock In {\em 2020 10th Annual Computing and Communication Workshop and
  Conference, CCWC 2020}, pages 640--646. Institute of Electrical and
  Electronics Engineers Inc., 1 2020.

\bibitem{Sahu2016SentimentAlgorithms}
Tirath~Prasad Sahu and Sanjeev Ahuja.
\newblock {Sentiment analysis of movie reviews: A study on feature selection
  and classification algorithms}.
\newblock In {\em International Conference on Microelectronics, Computing and
  Communication, MicroCom 2016}. Institute of Electrical and Electronics
  Engineers Inc., 7 2016.

\bibitem{Fauzi2019Word2VecLanguage}
M.~Ali Fauzi.
\newblock {Word2Vec model for sentiment analysis of product reviews in
  Indonesian language}.
\newblock {\em International Journal of Electrical and Computer Engineering
  (IJECE)}, 9(1):525, 2 2019.

\bibitem{Munikar2019Fine-grainedBERT}
Manish Munikar, Sushil Shakya, and Aakash Shrestha.
\newblock {Fine-grained Sentiment Classification using BERT}.
\newblock {\em International Conference on Artificial Intelligence for
  Transforming Business and Society, AITB 2019}, 10 2019.

\bibitem{Mansoor2020GlobalTime}
Muvazima Mansoor, Kirthika Gurumurthy, Anantharam~R U, and V~R~Badri Prasad.
\newblock {Global Sentiment Analysis Of COVID-19 Tweets Over Time}.
\newblock 10 2020.

\bibitem{Karimi2020AdversarialBERT}
Akbar Karimi, Leonardo Rossi, and Andrea Prati.
\newblock {Adversarial Training for Aspect-Based Sentiment Analysis with BERT}.
\newblock 1 2020.

\bibitem{Goodfellow2015ExplainingExamples}
Ian~J. Goodfellow, Jonathon Shlens, and Christian Szegedy.
\newblock {Explaining and harnessing adversarial examples}.
\newblock In {\em 3rd International Conference on Learning Representations,
  ICLR 2015 - Conference Track Proceedings}. International Conference on
  Learning Representations, ICLR, 12 2015.

\bibitem{TheExchanges}
{The Birth of Stock Exchanges}.

\bibitem{WhatStock}
{What Was the First Company to Issue Stock?}

\bibitem{Samuels2020SentimentResponses}
Antony Samuels and John Mcgonical.
\newblock {Sentiment Analysis on Customer Responses}.
\newblock 7 2020.

\bibitem{Pagolu2016SentimentMovements}
Venkata~Sasank Pagolu, Kamal Nayan~Reddy Challa, Ganapati Panda, and Babita
  Majhi.
\newblock {Sentiment Analysis of Twitter Data for Predicting Stock Market
  Movements}.
\newblock {\em International conference on Signal Processing}, page 2016, 10
  2016.

\bibitem{Medeiros2020TweetMarket}
Murilo~C. Medeiros and Vinicius R.~P. Borges.
\newblock {Tweet Sentiment Analysis Regarding the Brazilian Stock Market}.
\newblock pages 71--82. Sociedade Brasileira de Computacao - SB, 1 2020.

\bibitem{3Analysis}
{(3) (PDF) Examining the Effects of Pandemics on Stock Market Trends through
  Sentiment Analysis}.

\bibitem{Atkins2018FinancialPrice}
Adam Atkins, Mahesan Niranjan, and Enrico Gerding.
\newblock {Financial news predicts stock market volatility better than close
  price}.
\newblock {\em Journal of Finance and Data Science}, 4(2):120--137, 2018.

\bibitem{Vargas2017DeepArticles}
Manuel~R. Vargas, Beatriz~S.L.P. De~Lima, and Alexandre~G. Evsukoff.
\newblock {Deep learning for stock market prediction from financial news
  articles}.
\newblock {\em 2017 IEEE International Conference on Computational Intelligence
  and Virtual Environments for Measurement Systems and Applications, CIVEMSA
  2017 - Proceedings}, (June):60--65, 2017.

\bibitem{Vargas2018DeepArticles}
Manuel~R. Vargas, Carlos~E.M. Dos~Anjos, Gustavo~L.G. Bichara, and Alexandre~G.
  Evsukoff.
\newblock {Deep Leaming for Stock Market Prediction Using Technical Indicators
  and Financial News Articles}.
\newblock In {\em Proceedings of the International Joint Conference on Neural
  Networks}, volume 2018-July. Institute of Electrical and Electronics
  Engineers Inc., 10 2018.

\bibitem{Dang2016ImprovementArticles}
Minh Dang and Duc Duong.
\newblock {Improvement methods for stock market prediction using financial news
  articles}.
\newblock In {\em 2016 3rd National Foundation for Science and Technology
  Development Conference on Information and Computer Science (NICS)}, number
  May, pages 125--129. IEEE, 9 2016.

\bibitem{Shynkevich2015StockArticles}
Yauheniya Shynkevich, T.~M. McGinnity, Sonya Coleman, and Ammar Belatreche.
\newblock {Stock price prediction based on stock-specific and
  sub-industry-specific news articles}.
\newblock {\em Proceedings of the International Joint Conference on Neural
  Networks}, 2015-Septe, 2015.

\bibitem{Akita20162016Proceedings}
Ryo Akita.
\newblock {2016 IEEE/ACIS 15th International Conference on Computer and
  Information Science, ICIS 2016 - Proceedings}.
\newblock {\em 2016 IEEE/ACIS 15th International Conference on Computer and
  Information Science, ICIS 2016 - Proceedings}, 2016.

\bibitem{Kalyani2016StockAnalysis}
Joshi Kalyani, Prof. H.~N. Bharathi, and Prof.~Rao Jyothi.
\newblock {Stock trend prediction using news sentiment analysis}.
\newblock {\em International Journal of Computer Science and Information
  Technology}, 8(3):67--76, 7 2016.

\bibitem{Liu2018LeveragingNetwork}
Huicheng Liu.
\newblock {Leveraging Financial News for Stock Trend Prediction with
  Attention-Based Recurrent Neural Network}.
\newblock pages 1--24, 2018.

\bibitem{PDFModel}
{(PDF) A Chinese Stock Reviews Sentiment Analysis Based on BERT Model}.

\bibitem{Hajek2017CombiningReturns}
Petr H{\'{a}}jek.
\newblock {Combining bag-of-words and sentiment features of annual reports to
  predict abnormal stock returns}.
\newblock {\em Neural Computing and Applications 2017 29:7}, 29(7):343--358, 8
  2017.

\bibitem{Krauss2014PREDICTIVECASE}
Jonas~Sebastian Krauss and Detlef Schoder.
\newblock {PREDICTIVE ANALYTICS ON PUBLIC DATA – THE CASE}.
\newblock (June 2013), 2014.

\bibitem{Imbir2017PsychoevolutionaryPlutchik}
Kamil~K. Imbir.
\newblock {Psychoevolutionary Theory of Emotion (Plutchik)}.
\newblock In {\em Encyclopedia of Personality and Individual Differences},
  pages 1--9. Springer International Publishing, 2017.

\bibitem{IndiceWikipedia}
{{\'{I}}ndice Bovespa - Wikipedia}.

\bibitem{RandomTopics}
{Random Forest - an overview | ScienceDirect Topics}.

\bibitem{SupportScience}
{Support Vector Machine — Introduction to Machine Learning Algorithms | by
  Rohith Gandhi | Towards Data Science}.

\bibitem{Vosen2011ForecastingTrends}
Simeon Vosen and Torsten Schmidt.
\newblock {Forecasting private consumption: Survey-based indicators vs. Google
  trends}.
\newblock {\em Journal of Forecasting}, 30(6):565--578, 9 2011.

\bibitem{CHOI2012PredictingTrends}
HYUNYOUNG CHOI and HAL VARIAN.
\newblock {Predicting the Present with Google Trends}.
\newblock {\em Economic Record}, 88(SUPPL.1):2--9, 6 2012.

\bibitem{LatentWikipedia}
{Latent Dirichlet allocation - Wikipedia}.

\bibitem{NaiveScience}
{Naive Bayes Classifier. What is a classifier? | by Rohith Gandhi | Towards
  Data Science}.

\bibitem{DictionSimplified}
{Diction Software – Text Analysis Simplified}.

\bibitem{ResourcesDame}
{Resources // Textual Analysis // Software Repository for Accounting and
  Finance // University of Notre Dame}.

\bibitem{PDFData}
{(PDF) Comparative analysis of Stock Market Prediction Algorithms based on
  Twitter Data}.

\bibitem{Jampala2019PredictiveSensex}
Rajesh~C. Jampala, Prasanna~Kumar Goda, and Srinivasa~Rao Dokku.
\newblock {Predictive analytics in stock markets with special reference to BSE
  sensex}.
\newblock {\em International Journal of Innovative Technology and Exploring
  Engineering}, 8(6):615--619, 2019.

\bibitem{Li2020ANews}
Yang Li and Yi~Pan.
\newblock {A Novel Ensemble Deep Learning Model for Stock Prediction Based on
  Stock Prices and News}.
\newblock pages 1--15, 2020.

\bibitem{Guo2020ESG2Risk:Prediction}
Tian Guo.
\newblock {ESG2Risk: A Deep Learning Framework from ESG News to Stock
  Volatility Prediction}.
\newblock {\em SSRN Electronic Journal}, 2020.

\bibitem{AutomaticIs}
{Automatic document classification: what it is}.

\bibitem{Arras2017WhatApproach}
Leila Arras, Franziska Horn, Grégoire Montavon, Klaus~Robert M{\"{u}}ller, and
  Wojciech Samek.
\newblock {"What is relevant in a text document?": An interpretable machine
  learning approach}.
\newblock {\em PLoS ONE}, 12(8):1--19, 2017.

\bibitem{Gidofalvi2001UsingMovements}
Gyözö Gid{\'{o}}falvi.
\newblock {Using news articles to predict stock price movements}.
\newblock {\em Department of Computer Science and Engineering University of
  California San Diego}, page~9, 2001.

\bibitem{Akita2016DeepInformation}
Ryo Akita, Akira Yoshihara, Takashi Matsubara, and Kuniaki Uehara.
\newblock {Deep learning for stock prediction using numerical and textual
  information}.
\newblock In {\em 2016 IEEE/ACIS 15th International Conference on Computer and
  Information Science, ICIS 2016 - Proceedings}. Institute of Electrical and
  Electronics Engineers Inc., 8 2016.

\bibitem{Aamir2017StoryMarket}
Sarwar Aamir and Mazhar Qurat-ul ain.
\newblock {Story beneath story: Do magazine articles reveal forthcoming returns
  on stock market?}
\newblock {\em African Journal of Business Management}, 11(19):564--581, 10
  2017.

\bibitem{Kollintza-Kyriakoulia2018MeasuringTechniques}
Foteini Kollintza-Kyriakoulia, Manolis Maragoudakis, and Anastasia Krithara.
\newblock {Measuring the impact of financial news and social media on stock
  market modeling using time series mining techniques}.
\newblock {\em Algorithms}, 11(11), 2018.

\bibitem{Le2014DistributedDocuments}
Quoc~V. Le and Tomas Mikolov.
\newblock {Distributed Representations of Sentences and Documents}.
\newblock {\em 31st International Conference on Machine Learning, ICML 2014},
  4:2931--2939, 5 2014.

\bibitem{Ding2016Knowledge-DrivenPrediction}
Xiao Ding, Yue Zhang, T.~Liu, and Junwen Duan.
\newblock {Knowledge-Driven Event Embedding for Stock Prediction}.
\newblock {\em undefined}, 2016.

\bibitem{StemmingLemmatization}
{Stemming and lemmatization}.

\bibitem{DroppingWords}
{Dropping common terms: stop words}.

\bibitem{TokenizationKeras}
{Tokenization and Text Data Preparation with TensorFlow {\&} Keras}.

\bibitem{NLPMedium}
{NLP Guide: Identifying Part of Speech Tags using Conditional Random Fields |
  by Aiswarya Ramachandran | Analytics Vidhya | Medium}.

\bibitem{NLP:Science}
{NLP: Building Text Cleanup and PreProcessing Pipeline | by Dinesh Yadav |
  Towards Data Science}.

\bibitem{VADERTrading}
{VADER Sentiment Analysis in Algorithmic Trading}.

\bibitem{IntroductionScience}
{Introduction to Word Embedding and Word2Vec | by Dhruvil Karani | Towards Data
  Science}.

\bibitem{Kasthuriarachchy2014EnhancedAnalysis}
Buddhika~H. Kasthuriarachchy, Kasun De~Zoysa, and H.~L. Premaratne.
\newblock {Enhanced bag-of-words model for phrase-level sentiment analysis}.
\newblock {\em 2014 14th International Conference on Advances in ICT for
  Emerging Regions, ICTer 2014 - Conference Proceedings}, pages 210--214, 2014.

\bibitem{Qaiser2018TextDocuments}
Shahzad Qaiser and Ramsha Ali.
\newblock {Text Mining: Use of TF-IDF to Examine the Relevance of Words to
  Documents}.
\newblock {\em International Journal of Computer Applications}, 181(1):25--29,
  7 2018.

\bibitem{GoogleProcessing}
{Google AI Blog: Open Sourcing BERT: State-of-the-Art Pre-training for Natural
  Language Processing}.

\bibitem{Devlin2018BERT:Understanding}
Jacob Devlin, Ming-Wei Chang, Kenton Lee, and Kristina Toutanova.
\newblock {BERT: Pre-training of Deep Bidirectional Transformers for Language
  Understanding}.
\newblock {\em NAACL HLT 2019 - 2019 Conference of the North American Chapter
  of the Association for Computational Linguistics: Human Language Technologies
  - Proceedings of the Conference}, 1:4171--4186, 10 2018.

\bibitem{Pennington2014GloVe:Representation}
Jeffrey Pennington, Richard Socher, and Christopher~D. Manning.
\newblock {GloVe: Global vectors for word representation}.
\newblock In {\em EMNLP 2014 - 2014 Conference on Empirical Methods in Natural
  Language Processing, Proceedings of the Conference}, 2014.

\bibitem{Thakkar2021ADirections}
Ankit Thakkar and Kinjal Chaudhari.
\newblock {A comprehensive survey on deep neural networks for stock market: The
  need, challenges, and future directions}.
\newblock {\em Expert Systems with Applications}, 177:114800, 9 2021.

\bibitem{Petnehazi2018RecurrentForecasting}
Gábor Petneh{\'{a}}zi.
\newblock {Recurrent Neural Networks for Time Series Forecasting}.
\newblock 12 2018.

\bibitem{Pawar2019StockRNN}
Kriti Pawar, Raj~Srujan Jalem, and Vivek Tiwari.
\newblock {Stock Market Price Prediction Using LSTM RNN}.
\newblock {\em Advances in Intelligent Systems and Computing}, 841:493--503,
  2019.

\bibitem{PascanuOnNetworks}
Razvan Pascanu, Tomas Mikolov, and Yoshua Bengio.
\newblock {On the difficulty of training Recurrent Neural Networks}.

\bibitem{OVERCOMINGNETWORKS}
{OVERCOMING THE VANISHING GRADIENT PROBLEM IN PLAIN RECURRENT NETWORKS}.

\bibitem{BusinessHeadlines}
{Business News, Finance News, India News, BSE/NSE News, Stock Markets News,
  Sensex NIFTY, Latest Breaking News Headlines}.

\bibitem{FindKaggle}
{Find Open Datasets and Machine Learning Projects | Kaggle}.

\bibitem{EconomicData.world}
{Economic News Article Tone - dataset by crowdflower | data.world}.

\bibitem{YahooNews}
{Yahoo Finance – stock market live, quotes, business {\&} finance news}.

\bibitem{NSELtd.}
{NSE - National Stock Exchange of India Ltd.}

\bibitem{S-and-p-500-companies-financials/dataGitHub}
{s-and-p-500-companies-financials/data at master {\textperiodcentered}
  datasets/s-and-p-500-companies-financials {\textperiodcentered} GitHub}.

\bibitem{FTSEFinance}
{FTSE 100 ({\^{}}FTSE) charts, data {\&} news – Yahoo Finance}.

\bibitem{EURONEXTFinance}
{EURONEXT (EUXTF) Stock Price, News, Quote {\&} History - Yahoo Finance}.

\bibitem{SSEFinance}
{SSE 100 Index (000132.SS) Charts, Data {\&} News - Yahoo Finance}.

\bibitem{DAXFinance}
{DAX PERFORMANCE-INDEX ({\^{}}GDAXI) Historical Data - Yahoo Finance}.

\bibitem{NikkeiFinance}
{Nikkei 225 ({\^{}}N225) Charts, Data {\&} News - Yahoo Finance}.

\bibitem{GoogleNews-vectors-negative300.bin.gzDrive}
{GoogleNews-vectors-negative300.bin.gz - Google Drive}.

\bibitem{HomepageQuantPedia}
{Homepage - QuantPedia}.

\bibitem{WhatScience}
{What is Exploratory Data Analysis? | by Prasad Patil | Towards Data Science}.

\bibitem{Canova2017HowData}
Stefania Canova, Diego~Luigi Cortinovis, and Federico Ambrogi.
\newblock {How to describe univariate data}.
\newblock {\em Journal of Thoracic Disease}, 9(6):1741, 6 2017.

\bibitem{FAANGDefinition}
{FAANG Stocks Definition}.

\bibitem{NowMarketWatch}
{Now that Tesla has joined the S{\&}P 500, know these 3 things before investing
  - MarketWatch}.

\bibitem{Sandilands2014BivariateAnalysis}
Debra~(Dallie) Sandilands.
\newblock {Bivariate Analysis}.
\newblock {\em Encyclopedia of Quality of Life and Well-Being Research}, pages
  416--418, 2014.

\bibitem{StandingScience}
{Standing Out From the Cloud: How to Shape and Format a Word Cloud | by Andrew
  Jamieson | Towards Data Science}.

\bibitem{PDFDatab}
{(PDF) The box plot: A simple visual method to interpret data}.

\bibitem{FTSENews}
{FTSE 100 rallies amid Covid vaccine rollout - BBC News}.

\bibitem{WhySlowdown}
{Why Indian stock markets have hit all-time high despite Covid, lockdown,
  record slowdown}.

\bibitem{IndianHindu}
{Indian shares hit all-time high as country approves COVID-19 vaccines - The
  Hindu}.

\bibitem{PositivelyFinance}
{Positively Skewed Distribution - Overview and Applications in Finance}.

\bibitem{NormalProperties}
{Normal Distribution - Overview, Parameters, and Properties}.

\bibitem{AnScience}
{An Introduction to t-SNE with Python Example | by Andre Violante | Towards
  Data Science}.

\bibitem{OmidvarLearningHeadlines}
Amin Omidvar, Hossein Pourmodheji, Aijun An, and Gordon Edall.
\newblock {Learning to Determine the Quality of News Headlines}.

\bibitem{Yu2019AArchitectures}
Yong Yu, Xiaosheng Si, Changhua Hu, and Jianxun Zhang.
\newblock {A review of recurrent neural networks: Lstm cells and network
  architectures}.
\newblock {\em Neural Computation}, 31(7):1235--1270, 7 2019.

\bibitem{SherstinskyFundamentalsNetwork}
Alex Sherstinsky.
\newblock {Fundamentals of Recurrent Neural Network (RNN) and Long Short-Term
  Memory (LSTM) Network}.

\bibitem{VanHoudt2020AModel}
Greg Van~Houdt, Carlos Mosquera, and ·~Gonzalo N{\'{a}}poles.
\newblock {A review on the long short-term memory model}.
\newblock {\em Artificial Intelligence Review}, 53:5929--5955, 2020.

\bibitem{Sari2020TextFeatures}
Winda~Kurnia Sari, Dian~Palupi Rini, and Reza~Firsandaya Malik.
\newblock {Text Classification Using Long Short-Term Memory With GloVe
  Features}.
\newblock {\em Jurnal Ilmiah Teknik Elektro Komputer dan Informatika}, 5(2):85,
  2 2020.

\end{thebibliography}

\end{document}